\pdfoutput=1
\documentclass[11pt,a4paper]{article}
\usepackage[hyperref]{acl2019nopatch}
\usepackage{times}
\usepackage{latexsym}

\usepackage{url}

\aclfinalcopy % Uncomment this line for the final submission
\def\aclpaperid{723} %  Enter the acl Paper ID here

\usepackage{amsmath}
\usepackage{amsfonts}
\usepackage{xspace}
\usepackage{adjustbox}

\usepackage{ocr}
\usepackage{microtype}

\usepackage[vvarbb]{newtxmath}

\DeclareMathAlphabet{\mathcal}{OMS}{cmsy}{m}{n}

\usepackage{calc}

\usepackage{url}

\usepackage{subcaption}
\usepackage{multirow}
\usepackage{multicol}
\usepackage{booktabs}

\usepackage{tikz}
\usetikzlibrary{arrows,matrix,positioning,fit,calc,shapes,decorations.pathreplacing,patterns}
\pgfdeclarelayer{background}
\pgfdeclarelayer{backgroundoverlay}
\pgfsetlayers{background,backgroundoverlay,main}
\usepackage{pgfplots}
\pgfplotsset{compat=1.13}

\usepackage{nicefrac}
\usepackage{enumerate}

\newcommand{\NLL}{\operatorname{NLL}}

\newcommand{\xx}{\mathbf{x}}
\newcommand{\defn}[1]{\textbf{#1}}

\usepackage[capitalise]{cleveref}

\crefname{section}{\S}{\S\S}
\Crefname{section}{\S}{\S\S}
\crefname{table}{Table}{}
\crefname{figure}{Figure}{}
\crefname{footnote}{footnote}{footnotes}   
\crefname{equation}{equation}{equations}   
\crefformat{section}{\S#2#1#3}  % remove space between section symbol and the number

\title{What Kind of Language Is Hard to Language-Model?}

\author{Sabrina J. Mielke$^1$ \; Ryan Cotterell$^1$ \; Kyle Gorman$^2$$^,$$^3$ \; Brian Roark$^3$ \; Jason Eisner$^1$ \\
  ${}^1$ Department of Computer Science, Johns Hopkins University \\
  ${}^2$ Program in Linguistics, Graduate Center, City University of New York  \hspace*{0.3in}
  ${}^3$ Google \\
  {\tt \{sjmielke@,ryan.cotterell@\}jhu.edu} \;\; {\tt kgorman@gc.cuny.edu} \\ {\tt roark@google.com} \;\; {\tt jason@cs.jhu.edu} \\}

\date{}

\def\sentence{s}
\def\sentunits{n}
\def\langdifficulty{d}

\def\idlang{j}
\def\idsent{i}
\def\idlangsent{{\idsent\idlang}}

\begin{document}
\maketitle
\begin{abstract}
  How language-agnostic are current state-of-the-art NLP tools? Are there some types of language that are easier to model with current methods? In prior work \citep{CotMieEis18All} we attempted to address this question for language modeling, and observed that recurrent neural network language models do not perform equally well over all the high-resource European languages found in the Europarl corpus. We speculated that inflectional morphology may be the primary culprit for the discrepancy. In this paper, we extend these earlier experiments to cover 69 languages from 13 language families using a multilingual Bible corpus. Methodologically, we introduce a new paired-sample multiplicative mixed-effects model to obtain language difficulty coefficients from at-least-pairwise parallel corpora. In other words, the model is aware of inter-sentence variation and can handle missing data. Exploiting this model, we show that ``translationese'' is not any easier to model than natively written language in a fair comparison. Trying to answer the question of what features difficult languages have in common, we try and fail to reproduce our earlier \citep{CotMieEis18All} observation about morphological complexity and instead reveal far simpler statistics of the data that seem to drive complexity in a much larger sample.
\end{abstract}

\thispagestyle{plain}
\pagestyle{plain}

\section{Introduction}

\begin{figure}
  \begin{adjustbox}{width=\linewidth}
    \begin{tikzpicture}
      % ------------ LEFT HALF ------------
      \draw[step=1.0,black,thin] (0,-.2) grid (3,4);
      % Sentence 1
      \draw[fill=black] (0,3) rectangle (1,4);
      % \node[anchor=center,font=\ttfamily\bfseries,scale=0.35] at (0.5,3.5) {\begin{minipage}{4.2em}\color{white}Re\-sump\-tion of the session\end{minipage}};
      \draw[preaction={fill=black}, pattern=crosshatch dots, pattern color=black] (1,3) rectangle (2,4);
      % \node[anchor=center,font=\ttfamily\bfseries,scale=0.35] at (1.5,3.5) {\begin{minipage}{4.2em}\color{gray}Wie\-der\-auf\-nah\-me der ...\end{minipage}};
      % Sentence 2
      \draw[preaction={fill=black}, pattern=crosshatch dots, pattern color=black] (0,2) rectangle (1,3);
      % \node[anchor=center,font=\ttfamily\bfseries,scale=0.35] at (0.5,2.5) {\begin{minipage}{4.2em}\color{gray}The peace that ...\end{minipage}};
      \draw[fill=black] (1,2) rectangle (2,3);
      % \node[anchor=center,font=\ttfamily\bfseries,scale=0.35] at (1.5,2.5) {\begin{minipage}{4.2em}\color{white}Der gestern verein- ...\end{minipage}};
      \draw[preaction={fill=black}, pattern=crosshatch dots, pattern color=black] (2,2) rectangle (3,3);
      % \node[anchor=center,font=\ttfamily\bfseries,scale=0.35] at (2.5,2.5) {\begin{minipage}{4.2em}\color{gray}\textcyrillic{Мирът, който беше} ...\end{minipage}};
      % Sentence 3
      \draw[fill=black] (1,1) rectangle (2,2);
      % \node[anchor=center,font=\ttfamily\bfseries,scale=0.35] at (1.5,1.5) {\begin{minipage}{4.2em}\color{white}Obwohl wir nicht ...\end{minipage}};
      \draw[preaction={fill=black}, pattern=crosshatch dots, pattern color=black] (2,1) rectangle (3,2);
      % \node[anchor=center,font=\ttfamily\bfseries,scale=0.35] at (2.5,1.5) {\begin{minipage}{4.2em}\color{gray}\textcyrillic{Макар че не бяхме} ...\end{minipage}};
      % Sentence 4
      \draw[fill=black] (0,0) rectangle (1,1);
      % \node[anchor=center,font=\ttfamily\bfseries,scale=0.35] at (0.5,0.5) {\begin{minipage}{4.2em}\color{white}Now we can finally ...\end{minipage}};
      \draw[preaction={fill=black}, pattern=crosshatch dots, pattern color=black] (2,0) rectangle (3,1);
      % \node[anchor=center,font=\ttfamily\bfseries,scale=0.35] at (2.5,0.5) {\begin{minipage}{4.2em}\color{gray}\textcyrillic{Накрая всички можем} ...\end{minipage}};
      % Actual text inclusion cause rendering issues
      \node at (1.5,2) {\includegraphics{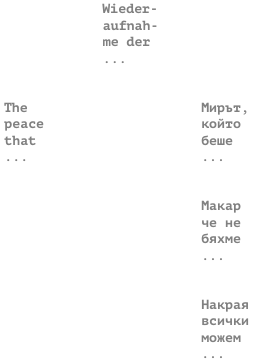}};
      % LANGUAGES header
      \node[anchor=south] at (0.5,4) {en};
      \node[anchor=south] at (1.5,4) {de};
      \node[anchor=south] at (2.5,4) {b\smash{g}};
      % INTENTS header
      \node[anchor=east] at (0,3.5) {1};
      \node[anchor=east] at (0,2.5) {2};
      \node[anchor=east] at (0,1.5) {3};
      \node[anchor=east] at (0,0.5) {4};
      % Bottom label
      \node[anchor=north] at (1.5,-.5) {aligned multi-text};

      % ------------ CONNECTOR ------------
      \draw[-stealth,double] (3.2,2) to node[above]{language} node[below]{model} (4.6,2);

      % ------------ RIGHT HALF ------------
      \begin{scope}[shift={(4.8,0)}]
        \draw[step=1.0,black,thin] (0,-.2) grid (3,4);
        % Sentence 1
        \draw[fill=yellow!99!red] (0,3) rectangle (1,4); \node at (0.5,3.5) {$y_{1,\text{en}}$};
        \draw[fill=yellow!89!red] (1,3) rectangle (2,4); \node at (1.5,3.5) {$y_{1,\text{de}}$};
        % Sentence 2
        \draw[fill=yellow!75!red] (0,2) rectangle (1,3); \node at (0.5,2.5) {$y_{2,\text{en}}$};
        \draw[fill=yellow!55!red] (1,2) rectangle (2,3); \node at (1.5,2.5) {$y_{2,\text{de}}$};
        \draw[fill=yellow!85!red] (2,2) rectangle (3,3); \node at (2.5,2.5) {$y_{2,\text{bg}}$};
        % Sentence 3
        \draw[fill=yellow!10!red] (1,1) rectangle (2,2); \node at (1.5,1.5) {$y_{3,\text{de}}$};
        \draw[fill=yellow!20!red] (2,1) rectangle (3,2); \node at (2.5,1.5) {$y_{3,\text{bg}}$};
        % Sentence 4
        \draw[fill=yellow!70!red] (0,0) rectangle (1,1); \node at (0.5,0.5) {$y_{4,\text{en}}$};
        \draw[fill=yellow!75!red] (2,0) rectangle (3,1); \node at (2.5,0.5) {$y_{4,\text{bg}}$};
        % HEADER
        \node[anchor=center,align=center] at (1.5,4.5) {\footnotesize Difficulty estimation\\[-.2em]\footnotesize from sentence surprisal};
        % \sentunits_\idsent
        \node[anchor=west, fill=yellow!97!red] at (3.4,3.5) {$\sentunits_1$};
        \node[anchor=west, fill=yellow!75!red] at (3.4,2.5) {$\sentunits_2$};
        \node[anchor=west, fill=yellow!20!red] at (3.4,1.5) {$\sentunits_3$};
        \node[anchor=west, fill=yellow!70!red] at (3.4,0.5) {$\sentunits_4$};
        \node[anchor=center] at (3.2,3.5) {$\Rightarrow$};
        \node[anchor=center] at (3.2,2.5) {$\Rightarrow$};
        \node[anchor=center] at (3.2,1.5) {$\Rightarrow$};
        \node[anchor=center] at (3.2,0.5) {$\Rightarrow$};
        % langdifficulty_idlang
        \node[anchor=north, fill=yellow!80!red] at (0.5,-.5) {$\langdifficulty_\text{en}$};
        \node[anchor=north, fill=yellow!60!red] at (1.5,-.5) {$\langdifficulty_\text{de}$};
        \node[anchor=north, fill=yellow!90!red] at (2.5,-.5) {$\langdifficulty_\text{bg}$};
        \node[anchor=center] at (0.5,-.2) {\rotatebox{270}{$\Rightarrow$}};
        \node[anchor=center] at (1.5,-.2) {\rotatebox{270}{$\Rightarrow$}};
        \node[anchor=center] at (2.5,-.2) {\rotatebox{270}{$\Rightarrow$}};
      \end{scope}
    \end{tikzpicture}
  \end{adjustbox}
  \caption{Jointly estimating the information $\sentunits_\idsent$ present in each multi-text intent $\idsent$ \emph{and} the difficulty $\langdifficulty_\idlang$ of each language $\idlang$. At left, gray text indicates translations of the original (white) sentence in the same row. At right, darker cells indicate higher surprisal/difficulty.  
    Empty cells indicate missing translations.
    English (en) is missing a hard sentence and Bulgarian (bg) is missing an easy sentence, but this does not mislead our method into estimating English as easier than Bulgarian.}
  \label{fig:sentencematrix}
\end{figure}

Do current NLP tools serve all languages? Technically, yes, as there are rarely hard constraints that prohibit application to specific languages, as long as
there is data annotated for the task. However, in practice, the answer is more nuanced: as most studies seem to (unfairly) assume English is representative of the world's languages \citep{Ben09Linguistically}, we do not have a clear idea
how well models perform cross-linguistically in a controlled setting. In this work, we look at current methods for language modeling
and attempt to determine whether there are typological properties that make
certain languages harder to language-model than others.\looseness=-1

One of the oldest tasks in NLP \citep{shannon1951prediction} is language modeling, which attempts to estimate a distribution $p(\xx)$ over strings $\xx$ of a language.
Recent years have seen impressive improvements with recurrent neural language models \citep[e.g.,][]{MerKesSoc18Analysis}.
Language modeling is an important component of tasks such as speech recognition, machine translation, and text normalization.  It has also enabled the construction of contextual word embeddings that provide impressive performance gains in many other NLP tasks \citep{PetNeuIyy18Deep}---though those downstream evaluations, too, have focused on a small number of (mostly English) datasets.\looseness=-1

In prior work \citep{CotMieEis18All}, we compared languages in terms of the difficulty of language modeling, controlling for differences in content by using a multi-lingual, fully parallel text corpus.  
Few such corpora exist: in that paper, we
made use of the Europarl corpus which, unfortunately, is not very typologically diverse.  Using a corpus with relatively few (and often related) languages limits the kinds of conclusions that can be drawn from any resulting comparisons.  In this paper, we present an alternative method that does not require the corpus to be {\em fully} parallel, so that collections consisting of many more languages can be compared.
Empirically, we report language-modeling results on 62 languages from 13 language families using Bible translations, and on the 21 languages used in the European Parliament proceedings.

We suppose that a language model's surprisal on a sentence---the negated log of the probability it assigns to the sentence---reflects not only the length and complexity of the specific sentence, but also the general difficulty that the model has in predicting sentences of that language.  Given language models of diverse languages, we jointly recover each language's difficulty parameter. Our regression formula explains the variance in the dataset better than previous approaches and can also deal with missing translations for some purposes.

Given these difficulty estimates, we conduct a correlational study, asking which typological features of a language are predictive of modeling difficulty. Our results suggest that simple properties of a language---the word inventory and (to a lesser extent) the raw character sequence length---are statistically significant indicators of modeling difficulty within our large set of languages.
In contrast, we fail to reproduce our earlier results from \citet{CotMieEis18All},\footnote{We can certainly \defn{replicate} those results in the sense that, using the surprisals from those experiments, we achieve the same correlations.  However, we did not \defn{reproduce} the results under new conditions \cite{drummond-2009}.  Our new conditions included a larger set of languages, a more sophisticated difficulty estimation method, and---perhaps crucially---improved language modeling families that tend to achieve better surprisals (or equivalently, better perplexity).} which suggested morphological complexity as an indicator of modeling complexity.  In fact, we find no tenable correlation to a wide variety of typological features, taken from the WALS dataset and other sources. Additionally, exploiting our model's ability to handle missing data, we directly test the hypothesis that translationese leads to easier language-modeling \citep{baker1993corpus,lembersky-ordan-wintner:2012:EACL2012}. We ultimatelycast doubt on this claim, showing that, under the strictest controls, translationese is \emph{different}, but not any \emph{easier} to model according to our notion of difficulty.

We conclude with a recommendation: The world being small, typology is in practice a small-data problem. there is a real danger that cross-linguistic studies will under-sample and thus over-extrapolate. We outline directions for future, more robust, investigations, and further caution that future work of this sort should focus on datasets with far more languages, something our new methods now allow.\looseness=-1

\section{The Surprisal of a Sentence}\label{sec:estimate-difficulty-by-lming}

When trying to estimate the difficulty (or complexity) of a language, we face a problem: the predictiveness of a language model on a domain of text will reflect not only the language that the text is written in, but also the topic, meaning, style, and information density of the text.
To measure the effect due only to the language, we would like to compare on datasets that are matched for the other variables, to the extent possible.  The datasets should all contain the same content, the only difference being the language in which it is expressed.  

\subsection{Multitext for a Fair Comparison}

To attempt a fair comparison, we make use of \defn{multitext}---sentence-aligned\footnote{\label{ft:verse}Both corpora we use align small paragraphs instead of sentences, but for simplicity we will call them ``sentences.''} translations of the \emph{same content} in multiple languages.
Different surprisals on the translations of the same sentence reflect quality differences in the language models, unless the translators added or removed information.%
\footnote{A translator might add or remove information out of helpfulness, sloppiness, showiness, consideration for their audience's background knowledge, or deference to the conventions of the target language.  For example, English conventions make it almost obligatory to express number (via morphological inflection), but make it optional to express evidentiality (e.g., via an explicit modal construction); other languages are different.\label{fn:translators-add-remove}}

In what follows, we will distinguish between the $\idsent$\textsuperscript{th} \defn{sentence} in language $\idlang$, which is a specific string $\sentence_\idlangsent$, and the $\idsent$\textsuperscript{th} \defn{intent}, the shared abstract thought that gave rise to all the sentences $\sentence_{\idsent 1},\sentence_{\idsent 2}, \ldots$. %, liberated from the manacles of punctuational arbitrarity.
For simplicity, suppose for now that we have a fully parallel corpus.  
We select, say, 80\% of the intents.\footnote{In practice, we use $\nicefrac{2}{3}$ of the raw data to train our models, $\nicefrac{1}{6}$ to tune them and the remaining $\nicefrac{1}{6}$ to test them.}  We use the English sentences that express these intents to train an English language model, and test it on the sentences that express the remaining 20\% of the intents.  We will later drop the assumption of a fully parallel corpus (\cref{sec:aggregation-regression}), which will help us to estimate 
the effects of translationese (\cref{sec:translationese-results}).\looseness=-1

\subsection{Comparing Surprisal Across Languages}

Given some test sentence $\sentence_\idlangsent$, a language model $p$ defines its \defn{surprisal}: the negative log-likelihood $\NLL(\sentence_\idlangsent) = -\log_2 p(\sentence_\idlangsent)$.  This can be interpreted as the number of bits needed to represent the sentence under a compression scheme that is derived from the language model, with high-probability sentences requiring the fewest bits.  Long or unusual sentences tend to have high surprisal---but high surprisal can also reflect a language's model's \emph{failure to anticipate} predictable words.  In fact, language models for the same language are often comparatively evaluated by their \emph{average surprisal} on a corpus (the \defn{cross-entropy}).  \newcite{CotMieEis18All} similarly compared language models for different languages, using a multitext corpus.

Concretely, recall that $\sentence_\idlangsent$ and $\sentence_{\idlangsent'}$ should contain, at least in principle, the same information for two languages $\idlang$ and $\idlang'$---they are translations of each other.
But, if we find that $\NLL(\sentence_\idlangsent) > \NLL(\sentence_{\idlangsent'})$, we must assume that either $\sentence_\idlangsent$ contains more information than $\sentence_{\idlangsent'}$, or that our language model was simply able to predict it less well.\footnote{The former might be the result of overt marking of, say, evidentiality or gender, which adds information. We hope that these differences are taken care of by diligent translators producing faithful translations in our multitext corpus.}
If we were to assume that our language models were perfect in the sense that they captured the true probability distribution of a language, we could make the former claim; but we suspect that much of the difference can be explained by our imperfect LMs rather than inherent differences in the expressed information (see the discussion in \cref{fn:translators-add-remove}).

\subsection{Our Language Models}
Specifically, the crude tools we use are recurrent neural network language models (RNNLMs) over different types of subword units.  For fairness, it is of utmost importance that these language models are \defn{open-vocabulary}, i.e., they predict the entire string and cannot cheat by predicting only \textsc{unk} (``unknown'') for some words of the language.\footnote{We restrict the set of characters to those that we see at least 25 times in the training set, replacing all others with a new symbol $\Diamond$, as is common and easily defensible in open-vocabulary language modeling \citep{MieEis18Spell}. We make an exception for Chinese, where we only require each character to appear at least twice.  These thresholds result in negligible ``out-of-alphabet'' rates for all languages.}

\paragraph{Char-RNNLM}
The first open-vocabulary RNNLM is the one of \citet{SutMarHin11Generating}, whose model generates a sentence, not word by word, but rather character by character.
An obvious drawback of the model is that it has no explicit representation of reusable substrings \citep{MieEis18Spell}, but the fact that it does not rely on a somewhat arbitrary word segmentation or tokenization makes it attractive for this study. We use a more current version based on LSTMs \citep{HocSch97Long}, using the implementation of \citet{MerKesSoc18Analysis} with the char-PTB parameters.

\begin{figure}
  \begin{adjustbox}{width=\linewidth}
    \begin{tikzpicture}
      \begin{axis}[
          width=25em,
          height=11em,
          ymax=6010000,
          legend style={at={(0.5, -.1)},anchor=north,draw=none,fill=none,column sep=.1em,name=legend},
          legend columns=7,
          legend cell align=left,
          mark options={scale=0.4}]
        % ALL VALUES
        \addplot[smooth,red,mark=text,text mark=bg] coordinates {(0.1, 5633551.21) (0.2, 5510954.80) (0.3, 5477450.68) (0.4, 5417901.43) (0.5, 5408816.61) (0.6, 5400489.11) (0.7, 5411222.35) (0.8, 5449696.01) (0.9, 5445833.68) (1.0, 5446188.27)};
        \addplot[smooth,blue,mark=text,text mark=cs] coordinates {(0.1, 5902016.77) (0.2, 5767526.82) (0.3, 5697313.30) (0.4, 5701967.07) (0.5, 5691711.74) (0.6, 5722710.07) (0.7, 5756528.79) (0.8, 5781379.76) (0.9, 5747815.81) (1.0, 5749168.52)};
        \addplot[smooth,green,mark=text,text mark=da] coordinates {(0.1, 5543016.74) (0.2, 5412079.67) (0.3, 5332888.00) (0.4, 5316404.07) (0.5, 5305808.11) (0.6, 5312718.27) (0.7, 5305956.99) (0.8, 5326711.90) (0.9, 5336336.58) (1.0, 5336010.19)};
        \addplot[smooth,yellow,mark=text,text mark=de] coordinates {(0.1, 6170041.86) (0.2, 5933509.27) (0.3, 5876357.09) (0.4, 5804515.57) (0.5, 5816482.10) (0.6, 5818846.53) (0.7, 5839229.43) (0.8, 5874563.86) (0.9, 5860283.35) (1.0, 5863856.20)};
        \addplot[smooth,brown,mark=text,text mark=el] coordinates {(0.1, 5953945.62) (0.2, 5780277.39) (0.3, 5696657.66) (0.4, 5643986.93) (0.5, 5641680.76) (0.6, 5647256.90) (0.7, 5633509.80) (0.8, 5641501.63) (0.9, 5671559.07) (1.0, 5667307.77)};
        \addplot[smooth,black,mark=text,text mark=en] coordinates {(0.1, 5770105.99) (0.2, 5579521.29) (0.3, 5523764.58) (0.4, 5467494.23) (0.5, 5417916.77) (0.6, 5413740.51) (0.7, 5400549.50) (0.8, 5395567.65) (0.9, 5426821.06) (1.0, 5400091.73)};
        \addplot[smooth,gray,mark=text,text mark=es] coordinates {(0.1, 5799711.01) (0.2, 5656045.59) (0.3, 5602344.54) (0.4, 5587582.12) (0.5, 5570624.65) (0.6, 5589247.32) (0.7, 5584445.34) (0.8, 5565694.38) (0.9, 5570645.38) (1.0, 5574092.47)};
        \addplot[smooth,pink,mark=text,text mark=et] coordinates {(0.1, 5742590.94) (0.2, 5616407.40) (0.3, 5634560.58) (0.4, 5638216.64) (0.5, 5662213.95) (0.6, 5673604.81) (0.7, 5716463.56) (0.8, 5716462.97) (0.9, 5712012.00) (1.0, 5716463.23)};
        \addplot[smooth,violet,mark=text,text mark=fi] coordinates {(0.1, 5612958.68) (0.2, 5513413.83) (0.3, 5471188.95) (0.4, 5522465.33) (0.5, 5509071.61) (0.6, 5582243.30) (0.7, 5617353.03) (0.8, 5617829.41) (0.9, 5621642.12) (1.0, 5622800.64)};
        \addplot[smooth,olive,mark=text,text mark=fr] coordinates {(0.1, 5727071.99) (0.2, 5553783.20) (0.3, 5560689.49) (0.4, 5481288.82) (0.5, 5444302.23) (0.6, 5451103.64) (0.7, 5467389.76) (0.8, 5475392.43) (0.9, 5470942.84) (1.0, 5472716.67)};
        \addplot[smooth,orange,mark=text,text mark=hu] coordinates {(0.1, 5969485.18) (0.2, 5875855.74) (0.3, 5856883.35) (0.4, 5821016.80) (0.5, 5879411.16) (0.6, 5890451.54) (0.7, 5955969.19) (0.8, 5955268.76) (0.9, 5956682.01) (1.0, 5955970.58)};
        \addplot[smooth,red,mark=text,text mark=it] coordinates {(0.1, 5792392.06) (0.2, 5668560.34) (0.3, 5633134.54) (0.4, 5564388.60) (0.5, 5626003.76) (0.6, 5571514.71) (0.7, 5540665.81) (0.8, 5545819.09) (0.9, 5529211.84) (1.0, 5529880.09)};
        \addplot[smooth,blue,mark=text,text mark=lt] coordinates {(0.1, 5576749.66) (0.2, 5527786.72) (0.3, 5474965.97) (0.4, 5515020.12) (0.5, 5534509.27) (0.6, 5546916.40) (0.7, 5558064.88) (0.8, 5558894.57) (0.9, 5558894.65) (1.0, 5560125.71)};
        \addplot[smooth,green,mark=text,text mark=lv] coordinates {(0.1, 5610881.24) (0.2, 5450563.03) (0.3, 5420911.03) (0.4, 5409310.27) (0.5, 5424061.52) (0.6, 5411226.65) (0.7, 5433930.86) (0.8, 5469517.76) (0.9, 5467211.26) (1.0, 5470763.66)};
        \addplot[smooth,yellow,mark=text,text mark=nl] coordinates {(0.1, 5964400.30) (0.2, 5751304.19) (0.3, 5682253.26) (0.4, 5653042.73) (0.5, 5667989.77) (0.6, 5662730.20) (0.7, 5626900.57) (0.8, 5639523.85) (0.9, 5630760.65) (1.0, 5631288.99)};
        \addplot[smooth,brown,mark=text,text mark=pl] coordinates {(0.1, 5949696.17) (0.2, 5779052.90) (0.3, 5766421.26) (0.4, 5731576.16) (0.5, 5713960.14) (0.6, 5737408.35) (0.7, 5743053.99) (0.8, 5776260.56) (0.9, 5792318.90) (1.0, 5798075.17)};
        \addplot[smooth,black,mark=text,text mark=pt] coordinates {(0.1, 5860685.97) (0.2, 5726676.75) (0.3, 5689059.88) (0.4, 5625655.63) (0.5, 5602223.97) (0.6, 5592952.00) (0.7, 5612627.57) (0.8, 5597815.63) (0.9, 5588278.15) (1.0, 5616269.53)};
        \addplot[smooth,gray,mark=text,text mark=ro] coordinates {(0.1, 5944129.38) (0.2, 5669547.33) (0.3, 5620485.58) (0.4, 5669950.35) (0.5, 5666448.08) (0.6, 5536476.53) (0.7, 5549381.74) (0.8, 5561579.27) (0.9, 5614941.53) (1.0, 5620097.70)};
        \addplot[smooth,pink,mark=text,text mark=sk] coordinates {(0.1, 5682351.17) (0.2, 5568455.27) (0.3, 5499164.38) (0.4, 5473904.01) (0.5, 5476884.48) (0.6, 5508287.29) (0.7, 5512198.49) (0.8, 5542571.19) (0.9, 5544792.06) (1.0, 5550186.17)};
        \addplot[smooth,violet,mark=text,text mark=sl] coordinates {(0.1, 5621261.72) (0.2, 5528493.85) (0.3, 5510046.83) (0.4, 5451171.73) (0.5, 5467286.46) (0.6, 5486178.18) (0.7, 5479198.78) (0.8, 5508242.70) (0.9, 5508080.46) (1.0, 5505952.65)};
        \addplot[smooth,olive,mark=text,text mark=sv] coordinates {(0.1, 5542497.74) (0.2, 5404296.65) (0.3, 5345947.07) (0.4, 5346461.33) (0.5, 5324169.42) (0.6, 5345180.05) (0.7, 5333329.94) (0.8, 5347790.83) (0.9, 5348055.28) (1.0, 5349987.83)};
        % \legend{bg,cs,da,de,el,en,es,et,fi,fr,hu,it,lt,lv,nl,pl,pt,ro,sk,sl,sv}
      \end{axis}
    \end{tikzpicture}
  \end{adjustbox}
  \mbox{}\\[-1.6em]
  \begin{adjustbox}{width=\linewidth}
    \hspace{.9em}
    \begin{tikzpicture}
      \begin{axis}[
          width=23.7em,
          height=6em,
          enlarge x limits=0.1,
          yticklabels={,,}
          xtick={0.2,0.4,0.6,0.8,1.0},
          xticklabels={,,},
          xmin=0.1,
          ymax=1.02]
        \addplot coordinates {(0.1, 1.04377646078801) (0.2, 1.01718761996065) (0.3, 1.00941490940923) (0.4, 1.00488501419594) (0.5, 1.00490836215253) (0.6, 1.0053343351206) (0.7, 1.00681283970745) (0.8, 1.00914211584995) (0.9, 1.00963112564805) (1.0, 1.00991579350536)};
      \end{axis}
    \end{tikzpicture}
  \end{adjustbox}
  \caption{Top: For each language, total NLL of the dev corpus varies with the number of BPE merges, which is expressed on the $x$-axis as a fraction of the number of observed word types $|\mathcal{V}|$.\textsuperscript{\ref{ft:langs}} Bottom: Averaging over all 21 languages motivates a global value of 0.4.}
  \label{fig:bpe-tuning}
\end{figure}
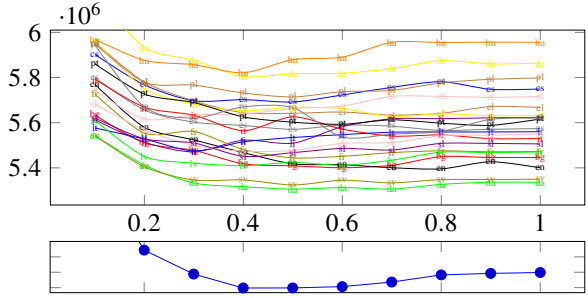

\paragraph{BPE-RNNLM}
BPE-based open-vocabulary language models make use of sub-word units instead of either words or characters and are a strong baseline on multiple languages \citep{MieEis18Spell}.  Before training the RNN, \defn{byte pair encoding} \citep[BPE;][]{SenHadBir16Neural} is applied globally to the training corpus, splitting each word (i.e., each space-separated substring) into one or more units.  The RNN is then trained over the sequence of units, which looks like this: ``The\ \ $|$ex$|$os$|$kel$|$eton\ \ $|$is\ \ $|$gener$|$ally\ \ $|$blue''.
The set of subword units is finite and determined from training data only, but it is a superset of the alphabet, making it possible to explain any novel word in held-out data via some segmentation.\footnote{In practice, in both training and testing, we only evaluate the probability of the canonical segmentation of the held-out string, rather than the total probability of all segmentations \citep[][Appendix D.2]{Kud18Subword,MieEis18Spell}.}
One important thing to note is that the size of this set can be tuned by specifying the number of BPE \defn{merges}, allowing us to smoothly vary between a word-level model ($\infty$ merges) and a kind of character-level model (0 merges).
As \cref{fig:bpe-tuning} shows, the number of merges that maximizes log-likelihood of our dev set differs from language to language.\footnote{\label{ft:langs}\cref{fig:bpe-tuning} shows the 21 languages of the Europarl dataset. Optimal values: 0.2 (et); 0.3 (fi, lt); 0.4 (de, es, hu, lv, sk,
  sl); 0.5 (da, fr, pl, sv); 0.6 (bg, ru); 0.7 (el); 0.8 (en); 0.9
  (it, pt).}
However, as we will see in \cref{fig:ep-difficulties-modeltuning}, tuning this parameter does not substantially influence our results.
We therefore will refer to the model with $0.4|\mathcal{V}|$ merges as BPE-RNNLM.

\section{Aggregating Sentence Surprisals}\label{sec:aggregation-regression}

\citet{CotMieEis18All} evaluated the model for language $\idlang$ simply
by its total surprisal $\sum_\idsent \NLL(\sentence_\idlangsent)$.  This comparative measure required a complete multitext corpus containing every sentence $\sentence_\idlangsent$ (the expression of the intent $\idsent$ in language $\idlang$).
We relax this requirement by using a fully probabilistic regression model that can deal with missing data (\cref{fig:sentencematrix}).%
\footnote{Specifically, we deal with data missing completely at random (MCAR), a strong assumption on the data generation process. More discussion on this can be found in \cref{app:missingness}.}
Our model predicts each sentence's surprisal $y_\idlangsent = \NLL(\sentence_\idlangsent)$ 
using an intent-specific ``information content'' factor  $\sentunits_\idsent$, which captures the inherent surprisal of the intent, combined with a language-specific difficulty factor $\langdifficulty_\idlang$.
This represents a better approach to varying sentence lengths and lets us work with missing translations in the test data (though it does not remedy our need for fully parallel language model training data).

\subsection{Model 1: Multiplicative Mixed-effects}

Model 1 is a multiplicative mixed-effects model:
\begin{align}
  y_\idlangsent &= \sentunits_\idsent \cdot \exp(\langdifficulty_\idlang ) \cdot \exp(\epsilon_\idlangsent) \label{eqn:y_n_d_e} \\
  \epsilon_\idlangsent &\sim \mathcal{N}(0,\sigma^2) \label{eqn:epsilon}
\end{align}
This says that each intent $\idsent$ has a latent \defn{size} of $\sentunits_\idsent$---measured in some abstract ``informational units''---%
that is observed indirectly in the various sentences $\sentence_\idlangsent$ that express the intent.  Larger $\sentunits_\idsent$ tend to yield longer
sentences.  
Sentence $\sentence_\idlangsent$ has $y_\idlangsent$ bits of surprisal; thus the multiplier $\nicefrac{y_\idlangsent}{\sentunits_\idsent}$ represents the number of bits that \emph{language $j$} used to express each informational unit of intent $i$, under our language model of language $j$.
Our mixed-effects model assumes that this multiplier is log-normally distributed over the sentences $i$: that is, $\log(\nicefrac{y_\idlangsent}{\sentunits_\idsent}) \sim \mathcal{N}(\langdifficulty_\idlang, \sigma^2)$, where mean $\langdifficulty_\idlang$ is the \defn{difficulty} of language $\idlang$.
That is, $\nicefrac{y_\idlangsent}{\sentunits_\idsent} = \exp(\langdifficulty_\idlang + \epsilon_\idlangsent)$ where $\epsilon_\idlangsent \sim \mathcal{N}(0,\sigma^2)$ is residual noise, yielding \crefrange{eqn:y_n_d_e}{eqn:epsilon}.%
\footnote{It is tempting to give each language its own $\sigma_\idlang^2$ parameter, but then the MAP estimate is pathological, since infinite likelihood can be attained by setting one language's $\sigma_\idlang^2$ to 0.}
We jointly fit the intent sizes $\sentunits_\idsent$ and the language difficulties $\langdifficulty_\idlang$.

\subsection{Model 2: Heteroscedasticity}\label{sec:model2}

Because it is multiplicative, Model 1 appropriately predicts that in each language $j$, intents with large $\sentunits_\idsent$ will not only have larger $y_\idlangsent$ values but these values will vary more widely. However, Model 1 is \defn{homoscedastic}: the variance $\sigma^2$ of $\log(\nicefrac{y_\idlangsent}{\sentunits_\idsent})$ is assumed to be independent of the independent variable $\sentunits_\idsent$, which predicts that the distribution of $y_\idlangsent$ should spread out \emph{linearly} as the information content $\sentunits_\idsent$ increases: e.g., $p(y_\idlangsent \geq 13 \mid \sentunits_\idsent=10) = p(y_\idlangsent \geq 26 \mid \sentunits_\idsent=20)$.
That assumption is questionable, since for a longer sentence, we would expect $\log \nicefrac{y_\idlangsent}{\sentunits_\idsent}$ to come closer to its mean $\langdifficulty_\idlang$ as the random effects of individual translational choices average out.\footnote{Similarly, flipping a fair coin 10 times results in 
 $5 \pm 1.58$ heads where 1.58 represents the standard deviation, but flipping it 20 times does not result in $10 \pm 1.58\cdot 2$ heads but rather $10 \pm 1.58 \cdot \sqrt{2}$ heads.  Thus, with more flips, the ratio heads/flips tends to fall closer to its mean 0.5.}
We address this issue by assuming that $y_\idlangsent$ results from $\sentunits_\idsent \in \mathbb{N}$ independent choices:
\begin{align}
  y_\idlangsent &= \exp(\langdifficulty_\idlang) \cdot \left( \sum_{k=1}^{\sentunits_\idsent} \exp \epsilon_{\idlangsent{}k} \right) \\ \label{eqn:heterosum}
  \epsilon_{\idlangsent{}k} &\sim \mathcal{N}(0,\sigma^2)
\end{align}
The number of bits for the $k$\textsuperscript{th} informational unit now varies by a factor of $\exp \epsilon_{\idlangsent{}k}$ that is log-normal and independent of the other units.
It is common to approximate the sum of independent log-normals by another log-normal distribution, matching mean and variance \citep[Fenton-Wilkinson approximation;][]{Fen60Sum},\footnote{There are better approximations, but even the only slightly more complicated Schwartz-Yeh approximation \citep{SchwartzYeh} already requires costly and complicated approximations in addition to lacking the generalizability to non-integral $\sentunits_\idsent$ values that we will obtain for the Fenton-Wilkinson approximation.}
yielding Model 2:
\edef\eqncntrcache{\theequation}
\begin{align}
  \setcounter{equation}{0}
  \color{gray} y_\idlangsent &\color{gray}= \sentunits_\idsent \cdot \exp(\langdifficulty_\idlang) \cdot \exp (\epsilon_\idlangsent)\\
  \setcounter{equation}{\eqncntrcache}
  \sigma_\idsent^2 &= \ln \Big( 1 + \tfrac{\exp(\sigma^2) - 1}{\sentunits_\idsent} \Big) \label{eqn:sigma_i} \\
  \epsilon_\idlangsent &\sim \mathcal{N}\Big( \tfrac{\sigma^2 - \sigma_\idsent^2}{2}\,,\; \sigma_\idsent^2 \Big), \label{eqn:fw}
\end{align}
in which the noise term $\epsilon_\idlangsent$ now depends on $n_\idsent$.  Unlike \eqref{eqn:heterosum}, this formula no longer
requires $\sentunits_\idsent \in \mathbb{N}$; we allow any $\sentunits_\idsent \in \mathbb{R}_{> 0}$, which will also let us use gradient descent in estimating $\sentunits_\idsent$.

In effect, fitting the model chooses each $\sentunits_\idsent$ so that the resulting intent-specific but language-independent distribution of $\sentunits_\idsent \cdot \exp(\epsilon_\idlangsent)$ values,\footnote{The distribution of $\epsilon_\idlangsent$ is the same for every $\idlang$.  It no longer has mean 0, but it depends only on $\sentunits_\idsent$.} after it is scaled by $\exp(\langdifficulty_\idlang)$ for each language $\idlang$, will assign high probability to the observed $y_\idlangsent$.  Notice that in Model 2, the scale of $\sentunits_\idsent$ becomes meaningful: fitting the model will choose the size of the abstract informational units so as to predict how rapidly $\sigma_\idsent$ falls off with $n_\idsent$.  This contrasts with Model 1, where doubling all the $\sentunits_\idsent$ values could be compensated for by halving all the $\exp(\langdifficulty_\idlang)$ values.

\subsection{Model 2L: An Outlier-Resistant Variant}

One way to make Model 2 more outlier-resistant is to use a Laplace distribution\footnote{One could also use a Cauchy distribution instead of the Laplace distribution to get even heavier tails, but we saw little difference between the two in practice.} instead of a Gaussian in \eqref{eqn:fw} as an approximation to the distribution of $\epsilon_\idlangsent$.  The Laplace distribution is heavy-tailed, so it is more tolerant of large residuals.
We choose its mean and variance just as in \eqref{eqn:fw}.
This heavy-tailed $\epsilon_{ij}$ distribution can be viewed as approximating a version of Model 2 in which the $\epsilon_{\idlangsent{}k}$ themselves follow some heavy-tailed distribution.

\subsection{Estimating model parameters}

We fit each regression model's parameters by L-BFGS.  We then evaluate the model's fitness by measuring its held-out data likelihood---that is, the probability it assigns to the $y_\idlangsent$ values for held-out intents $\idsent$.  Here we use the previously fitted $\langdifficulty_\idlang$ and $\sigma$ parameters, but we must newly fit $\sentunits_\idsent$ values for the new $\idsent$ using MAP estimates or posterior means.
A full comparison of our models under various conditions can be found in \cref{app:goodness-of-fit-plot}.  The primary findings are as follows.
On Europarl data (which has fewer languages), Model 2 performs best.  On the Bible corpora, all models are relatively close to one another, though the robust Model 2L gets more consistent results than Model 2 across data subsets.  We use MAP estimates under Model 2 for all remaining experiments for speed and simplicity.%
\footnote{Further enhancements are possible: we discuss our ``Model 3'' in \cref{app:model-3}, but it did not seem to fit better.}

\subsection{A Note on Bayesian Inference}

As our model of $y_{ij}$ values is fully generative, one could place priors on our parameters and do full inference of the posterior rather than performing MAP inference.
We did experiment with priors but found them so quickly overruled by the data  that it did not make much sense to spend time on them.

Specifically, for full inference, we implemented all models in STAN \citep{STAN}, a toolkit for fast, state-of-the-art inference using Hamiltonian Monte Carlo (HMC) estimation.
Running HMC unfortunately scales sublinearly with the number of sentences (and thus results in very long sampling times), and the posteriors we obtained were unimodal with relatively small variances (see also \cref{app:goodness-of-fit-plot}). We therefore work with the MAP estimates in the rest of this paper.

\section{The Difficulties of 69 languages}\label{sec:difficulty-results}

Having outlined our method for estimating language difficulty scores $\langdifficulty_\idlang$, we now seek data to do so for all our languages. If we wanted to cover the most languages possible with parallel text, we should surely look at the Universal Declaration of Human Rights, which has been translated into over 500 languages.
Yet this short document is far too small to train state-of-the-art language models.
In this paper, we will therefore follow previous work in using the Europarl corpus \citep{koehn2005europarl}, but also for the first time make use of 106 Bibles from \citet{MayCys14Creating}'s corpus.\looseness=-1

Although our regression models of the surprisals $y_\idlangsent$ can be estimated from incomplete multitext, the surprisals themselves are derived from the language models we are comparing.  To ensure that the language models are comparable, we want to train them on completely parallel data in the various languages.  For this, we seek complete multitext.

\subsection{Europarl: 21 Languages}
The Europarl corpus \citep{koehn2005europarl} contains decades worth of discussions of the European Parliament, where each intent appears in up to 21 languages.
It was previously used by \citet{CotMieEis18All} for its size and stability.  In \cref{sec:translationese-results}, we will also exploit the fact that each intent's original language is known.  To simplify our access to this information, we will use the ``Corrected \& Structured Europarl Corpus'' (CoStEP) corpus \citep{GraBatVol14Cleaning}.
From it, we extract the intents that appear in all 21 languages, as enumerated in \cref{ft:langs}.
The full extraction process and corpus statistics are detailed in \cref{app:europarl-extraction}.

\subsection{The Bible: 62 Languages}\label{sec:bible-selection}
The Bible is a religious text that has been used for decades as a dataset for massively multilingual NLP \citep{Resnik1999,yarowsky2001inducing,agic2016multilingual}.
Concretely, we use the tokenized\footnote{The fact that the resource is tokenized is (yet) another possible confound for this study: we are not comparing performance on languages, but on languages/Bibles \emph{with some specific translator and tokenization}.  It is possible that our $y_{ij}$ values for each language $j$ depend to a small degree on the tokenizer that was chosen for that language.}
and aligned collection assembled by \citet{MayCys14Creating}.
We use the smallest annotated subdivision (a single \emph{verse}) as a sentence in our difficulty estimation model; see \cref{ft:verse}.

Some of the Bibles in the dataset are incomplete.  As the Bibles include different sets of verses (intents), we have to select a set of Bibles that overlap strongly, so we can use the verses shared by all these Bibles to comparably train all our language models (and fairly test them: see \cref{app:missingness}).  We cast this selection problem as an integer linear program (ILP), which we solve exactly in a few hours using the Gurobi solver (more details on this selection in \cref{app:bible-selection}).
This optimal solution keeps 25996 verses, each of which appears across 106 Bibles in 62 languages,\footnote{afr, aln, arb, arz, ayr, bba, ben, bqc, bul, cac, cak, ceb, ces, cmn, cnh, cym, dan, deu, ell, eng, epo, fin, fra, guj, gur, hat, hrv, hun, ind, ita, kek, kjb, lat, lit, mah, mam, mri, mya, nld, nor, plt, poh, por, qub, quh, quy, quz, ron, rus, som, tbz, tcw, tgl, tlh, tpi, tpm, ukr, vie, wal, wbm, xho, zom} spanning 13 language families.%
\footnote{22 Indo-European, 6 Niger-Congo, 6 Mayan, 6 Austronesian, 4 Sino-Tibetan, 4 Quechuan, 4 Afro-Asiatic, 2 Uralic, 2 Creoles, 2 Constructed languages, 2 Austro-Asiatic, 1 Totonacan, 1 Aymaran.  For each language, we are reporting here the first family listed by Ethnologue \citep{paul2009ethnologue}, manually fixing tlh $\mapsto$ Constructed language.
It is unfortunate not to have more families or more languages per family. A broader sample could be obtained by taking only the New Testament---but unfortunately that has $< 8000$ verses, a meager third of our dataset that is already smaller that the usually considered tiny PTB dataset (see details in \cref{app:bible-selection}).}
We allow $j$ to range over the 106 Bibles, so when a language has multiple Bibles, we estimate a separate difficulty $d_j$ for each one.

\subsection{Results}

\begin{figure}
  \centering
  \begin{adjustbox}{width=0.98\linewidth}
    \begin{tikzpicture}
      \node at (-5, 0) {};
      \node at (-5, -0.3) {chars};
      \node at (-5, -1.5) {BPE (0.4$|\mathcal{V}|$)};
      \node at (-5, -2.7) {BPE (best per language)};
      \node[rotate=90] at (-0.5, .2) {\footnotesize\color{black!39} bg};
      \node[rotate=90] at (-0.05, .2) {\footnotesize\color{black!55} cs};
      \node[rotate=90] at (-1.2, .2) {\footnotesize\color{black!46} da};
      \node[rotate=90] at (5.692, .2) {\footnotesize\color{black!100} de};
      \node[rotate=90] at (-0.95, .2) {\footnotesize\color{black!68} el};
      \node[rotate=90] at (-2.4, .2) {\footnotesize\color{black!65} en};
      \node[rotate=90] at (1.135, .2) {\footnotesize\color{black!39} es};
      \node[rotate=90] at (1.439, .2) {\footnotesize\color{black!47} et};
      \node[rotate=90] at (-0.25, .2) {\footnotesize\color{black!42} fi};
      \node[rotate=90] at (2.4, .2) {\footnotesize\color{black!96} fr};
      \node[rotate=90] at (2.1, .2) {\footnotesize\color{black!27} hu};
      \node[rotate=90] at (-0.7, .2) {\footnotesize\color{black!41} it};
      \node[rotate=90] at (-1.5, .2) {\footnotesize\color{black!36} lt};
      \node[rotate=90] at (-3.520, .2) {\footnotesize\color{black!74} lv};
      \node[rotate=90] at (0.5, .2) {\footnotesize\color{black!27} nl};
      \node[rotate=90] at (3.085, .2) {\footnotesize\color{black!71} pl};
      \node[rotate=90] at (0.802, .2) {\footnotesize\color{black!22} pt};
      \node[rotate=90] at (0.15, .2) {\footnotesize\color{black!48} ro};
      \node[rotate=90] at (-2.2, .2) {\footnotesize\color{black!60} sk};
      \node[rotate=90] at (-2.65, .2) {\footnotesize\color{black!68} sl};
      \node[rotate=90] at (-1.95, .2) {\footnotesize\color{black!21} sv};
      \draw[draw=black!39,thick] plot [smooth,tension=0.3] coordinates {(-0.377955239488148, 0) (-0.9739251063486887, -1.4) (-1.1158481720194902, -2.8)};
      \draw[draw=black!55,thick] plot [smooth,tension=0.3] coordinates {(-0.1323085926233425, 0) (1.0278516055500786, -1.4) (1.1887209826369038, -2.8)};
      \draw[draw=black!46,thick] plot [smooth,tension=0.3] coordinates {(-1.0658241442941196, 0) (-2.0465219668547308, -1.4) (-1.9265367933694433, -2.8)};
      \draw[draw=black!100,thick] plot [smooth,tension=0.3] coordinates {(5.692874646187962, 0) (2.305755685787425, -1.4) (2.31439409741101, -2.8)};
      \draw[draw=black!68,thick] plot [smooth,tension=0.3] coordinates {(-0.6984299592906251, 0) (0.9865526541864789, -1.4) (1.0677750185816737, -2.8)};
      \draw[draw=black!65,thick] plot [smooth,tension=0.3] coordinates {(-2.3332483367941634, 0) (-0.5138131206928798, -1.4) (-1.1400715540303197, -2.8)};
      \draw[draw=black!39,thick] plot [smooth,tension=0.3] coordinates {(1.135681875288408, 0) (0.3293148943568358, -1.4) (0.6956254116102123, -2.8)};
      \draw[draw=black!47,thick] plot [smooth,tension=0.3] coordinates {(1.4393639043253081, 0) (0.418791474879987, -1.4) (0.517097506597386, -2.8)};
      \draw[draw=black!42,thick] plot [smooth,tension=0.3] coordinates {(-0.24469148300596544, 0) (-1.0566507309003992, -1.4) (-1.0176690352934226, -2.8)};
      \draw[draw=black!96,thick] plot [smooth,tension=0.3] coordinates {(2.272493698914957, 0) (-0.25622429213196796, -1.4) (-0.6045447520376779, -2.8)};
      \draw[draw=black!27,thick] plot [smooth,tension=0.3] coordinates {(2.2232830324171227, 0) (1.9357395814139755, -1.4) (2.1798811932655315, -2.8)};
      \draw[draw=black!41,thick] plot [smooth,tension=0.3] coordinates {(-0.620207108335713, 0) (0.16054606166526764, -1.4) (0.11873965907747142, -2.8)};
      \draw[draw=black!36,thick] plot [smooth,tension=0.3] coordinates {(-1.4679422070089376, 0) (-0.9010349992027489, -1.4) (-0.8596849910651405, -2.8)};
      \draw[draw=black!74,thick] plot [smooth,tension=0.3] coordinates {(-3.52042432405319, 0) (-1.7290406355795207, -1.4) (-1.4717547345744486, -2.8)};
      \draw[draw=black!27,thick] plot [smooth,tension=0.3] coordinates {(0.6052267738136718, 0) (0.8840076694248578, -1.4) (0.8424600793552894, -2.8)};
      \draw[draw=black!71,thick] plot [smooth,tension=0.3] coordinates {(3.0854468519043943, 0) (1.2610639866325268, -1.4) (1.2790510817958234, -2.8)};
      \draw[draw=black!22,thick] plot [smooth,tension=0.3] coordinates {(0.8022999534776476, 0) (0.6996570093216414, -1.4) (0.7011776809301429, -2.8)};
      \draw[draw=black!48,thick] plot [smooth,tension=0.3] coordinates {(-0.04675102816026899, 0) (1.0761627405617147, -1.4) (0.7198132122723577, -2.8)};
      \draw[draw=black!60,thick] plot [smooth,tension=0.3] coordinates {(-2.175117236076951, 0) (-0.8135262110861685, -1.4) (-0.655664317799115, -2.8)};
      \draw[draw=black!68,thick] plot [smooth,tension=0.3] coordinates {(-2.511321036524822, 0) (-0.7765072783941562, -1.4) (-0.8500806008349737, -2.8)};
      \draw[draw=black!21,thick] plot [smooth,tension=0.3] coordinates {(-2.062450040673347, 0) (-2.018199022589584, -1.4) (-1.982880972509854, -2.8)};
    \end{tikzpicture}
  \end{adjustbox}
  \caption{The Europarl language difficulties appear more similar, and are ordered differently, when the RNN models use BPE units instead of character units.  Tuning BPE per-language has a small additional effect.}
  \label{fig:ep-difficulties-modeltuning}
  \vspace{-\baselineskip}
\end{figure}
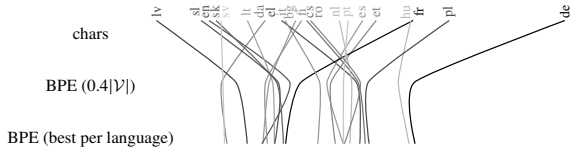

\begin{figure*}
  \centering
  \scalebox{.58}{
    \begin{tikzpicture}
      \begin{axis}[
        title={\Large Difficulties on Europarl},
        xlabel={\color{gray!40} difficulty ($\times 100$) using BPE-RNNLM with 0.4$|\mathcal{V}|$ merges},
        ylabel={\color{gray!40} difficulty ($\times 100$) using char-RNNLM},
        width=30em,
        height=30em,
        enlarge x limits=0,
        enlarge y limits=0,
        legend style={at={(0.5, -.1)},anchor=north,draw=none,fill=none,column sep=.1em,name=legend},
        legend columns=7,
        legend cell align=left,
        mark options={scale=0.8}]
        \addplot[mark=text, gray!20, text mark={\Huge easier with BPE}] coordinates {(-1.75, 9.5)};
        \addplot[mark=text, gray!20, text mark={\Huge easier with chars}] coordinates {(2, -5.5)};
        \addplot[mark=none, black!10] coordinates {(-4.7,-4.5) (5,5)};
        \addplot[mark=none, black!0] coordinates {(-4.7,-8) (5,11.5)};
        \addplot[blue!60, mark=text, text mark=bg] coordinates {(-1.9670708167404, -0.7587819619502)};
        \addplot[red!60, mark=text, text mark=cs] coordinates {(2.0348588145176, -0.2649679153851)};
        \addplot[blue!60, mark=text, text mark=da] coordinates {(-4.179167214849, -2.1546960304408)};
        \addplot[blue!60, mark=text, text mark=de] coordinates {(4.5083410941013, 10.7829209555326)};
        \addplot[red!60, mark=text, text mark=el] coordinates {(1.9538919080901, -1.4067078219063)};
        \addplot[red!60, mark=text, text mark=en] coordinates {(-1.0329427739309, -4.778888069376)};
        \addplot[blue!60, mark=text, text mark=es] coordinates {(0.6564702995606, 2.2459523546602)};
        \addplot[blue!60, mark=text, text mark=et] coordinates {(0.8340946882919, 2.8380708645704)};
        \addplot[blue!60, mark=text, text mark=fi] coordinates {(-2.1359513525203, -0.4905843656819)};
        \addplot[blue!60, mark=text, text mark=fr] coordinates {(-0.5137661050266, 4.4447294192267)};
        \addplot[blue!60, mark=text, text mark=hu] coordinates {(3.7984171670383, 4.3505425120136)};
        \addplot[red!60, mark=text, text mark=it] coordinates {(0.3205777234098, -1.2481715694386)};
        \addplot[red!60, mark=text, text mark=lt] coordinates {(-1.8185050263343, -2.9798440416863)};
        \addplot[red!60, mark=text, text mark=lv] coordinates {(-3.5192880991151, -7.3010022101144)};
        \addplot[red!60, mark=text, text mark=nl] coordinates {(1.7525677589509, 1.2031861455789)};
        \addplot[blue!60, mark=text, text mark=pl] coordinates {(2.4908471957063, 5.9879814669936)};
        \addplot[blue!60, mark=text, text mark=pt] coordinates {(1.3896140043971, 1.591862280618)};
        \addplot[red!60, mark=text, text mark=ro] coordinates {(2.1294900372492, -0.0935457967608)};
        \addplot[red!60, mark=text, text mark=sk] coordinates {(-1.6404342717693, -4.4476941460337)};
        \addplot[red!60, mark=text, text mark=sl] coordinates {(-1.5652001552448, -5.1531660405522)};
        \addplot[red!60, mark=text, text mark=sv] coordinates {(-4.1201212718338, -4.2123884124908)};
      \end{axis}
    \end{tikzpicture}
  }
  \hspace{.5em}
  \scalebox{.58}{
    \raisebox{2em}{
      \begin{tikzpicture}[yscale=60,xscale=.5]
        \node at (1.25, 0.101) {\Large vs.};
        \begin{scope}[opacity=0.8]
          \draw[-latex, gray!40] (1.27, 0.07) to node[above,rotate=270]{harder} (1.27, 0.09);
          \draw[-latex, gray!40] (1.23, -0.06) to node[above,rotate=90]{easier} (1.23, -0.08);
          \node[align=right, text width=1.1em] at (-.3, -0.0155) {bg};  % original: -0.013629263893453
          \node[align=right, text width=1.1em] at (-.3, 0.0055) {cs};  % original: 0.008849454495662
          \node[align=right, text width=1.1em] at (-.3, -0.033) {da};  % original: -0.031669316226449
          \node[align=right, text width=1.1em] at (-.3, 0.07645631024817) {de};  % original: 0.07645631024817
          \node[align=right, text width=1.1em] at (-.3, 0.001) {el};  % original: 0.002735920430919
          \node[align=right, text width=1.1em] at (-.3, -0.029059154216535) {en};  % original: -0.029059154216535
          \node[align=right, text width=1.1em] at (-.3, -0.0105) {fi};  % original: -0.013132678591011
          \node[align=right, text width=1.1em] at (-.3, 0.025) {fr};  % original: 0.019654816571001
          \node[align=right, text width=1.1em] at (-.3, 0.040744798395259) {hu};  % original: 0.040744798395259
          \node[align=right, text width=1.1em] at (-.3, -0.004637969230144) {it};  % original: -0.004637969230144
          \node[align=right, text width=1.1em] at (-.3, -0.023991745340103) {lt};  % original: -0.023991745340103
          \node[align=right, text width=1.1em] at (-.3, 0.014778769522649) {nl};  % original: 0.014778769522649
          \node[align=right, text width=1.1em] at (-.3, 0.019) {pt};  % original: 0.014907381425075
          \node[align=right, text width=1.1em] at (-.3, 0.01) {ro};  % original: 0.010179721202442
          \draw[draw,thick,densely dotted] (0.15,-0.013629263893453) to[out=0,in=180,looseness=0.8] (2.4,-0.018434183859483);
          \draw[draw,thick,densely dotted] (0.15,0.008849454495662) to[out=0,in=180,looseness=0.8] (2.4,0.045691777703497);
          \draw[draw,thick,densely dotted] (0.15,-0.031669316226449) to[out=0,in=180,looseness=0.8] (2.4,-0.008013942385278);
          \draw[draw,thick,densely dotted] (0.15,0.07645631024817) to[out=0,in=180,looseness=0.8] (2.4,0.045922004180027);
          \draw[draw,thick,densely dotted] (0.15,0.002735920430919) to[out=0,in=180,looseness=0.8] (2.4,-0.008651886319065);
          \draw[draw,thick,densely dotted] (0.15,-0.029059154216535) to[out=0,in=180,looseness=0.8] (2.4,-0.057537812610137);
          \draw[draw,thick,densely dotted] (0.15,-0.013132678591011) to[out=0,in=180,looseness=0.8] (2.4,0.023137162384962);
          \draw[draw,thick,densely dotted] (0.15,0.019654816571001) to[out=0,in=180,looseness=0.8] (2.4,-0.024595131063284);
          \draw[draw,thick,densely dotted] (0.15,0.040744798395259) to[out=0,in=180,looseness=0.8] (2.4,0.059378262621448);
          \draw[draw,thick,densely dotted] (0.15,-0.004637969230144) to[out=0,in=180,looseness=0.8] (2.4,0.022189439496381);
          \draw[draw,thick,densely dotted] (0.15,-0.023991745340103) to[out=0,in=180,looseness=0.8] (2.4,-0.07674329642782);
          \draw[draw,thick,densely dotted] (0.15,0.014778769522649) to[out=0,in=180,looseness=0.8] (2.4,-0.004943859145901);
          \draw[draw,thick,densely dotted] (0.15,0.014907381425075) to[out=0,in=180,looseness=0.8] (2.4,-0.008193781056251);
          \draw[draw,thick,densely dotted] (0.15,0.010179721202442) to[out=0,in=180,looseness=0.8] (2.4,-0.027743109117142);
          \node[align=left, text width=1.1em] at (2.9, -0.019) {bul};  % original: -0.018434183859483
          \node[align=left, text width=1.1em] at (2.9, 0.044) {ces};  % original: 0.045691777703497
          \node[align=left, text width=1.1em] at (2.9, -0.0035) {dan};  % original: -0.008013942385278
          \node[align=left, text width=1.1em] at (2.9, 0.049) {deu};  % original: 0.045922004180027
          \node[align=left, text width=1.1em] at (2.9, -0.0135) {ell};  % original: -0.008651886319065
          \node[align=left, text width=1.1em] at (2.9, -0.057537812610137) {eng};  % original: -0.057537812610137
          \node[align=left, text width=1.1em] at (2.9, 0.026) {fin};  % original: 0.023137162384962
          \node[align=left, text width=1.1em] at (2.9, -0.0245) {fra};  % original: -0.024595131063284
          \node[align=left, text width=1.1em] at (2.9, 0.059378262621448) {hun};  % original: 0.059378262621448
          \node[align=left, text width=1.1em] at (2.9, 0.021) {ita};  % original: 0.022189439496381
          \node[align=left, text width=1.1em] at (2.9, -0.07674329642782) {lit};  % original: -0.07674329642782
          \node[align=left, text width=1.1em] at (2.9, 0.002) {nld};  % original: -0.004943859145901
          \node[align=left, text width=1.1em] at (2.9, -0.0095) {por};  % original: -0.008193781056251
          \node[align=left, text width=1.1em] at (2.9, -0.0305) {ron};  % original: -0.027743109117142
        \end{scope}
      \end{tikzpicture}
    }
  }
  \hspace{.5em}
  \scalebox{.58}{
    \begin{tikzpicture}
      \begin{axis}[
          title={\Large \phantom{p}Difficulties on Bibles\phantom{p}},
          xlabel={\color{gray!40} difficulty ($\times 100$) using BPE-RNNLM with 0.4$|\mathcal{V}|$ merges},
          ylabel={\color{gray!40} difficulty ($\times 100$) using char-RNNLM},
          width=30em,
          height=30em,
          enlarge x limits=0,
          enlarge y limits=0,
          legend style={at={(0.5, -.1)},anchor=north,draw=none,fill=none,column sep=.1em,name=legend},
          legend columns=7,
          legend cell align=left,
          mark options={scale=0.8}]
          \addplot[mark=text, gray!20, text mark={\Huge easier with BPE}] coordinates {(-5, 17)};
          \addplot[mark=text, gray!20, text mark={\Huge easier with chars}] coordinates {(14, -13)};
          \addplot[mark=none, black!10] coordinates {(-19,-19) (21,21)};
          \addplot[mark=none, black!0] coordinates {(-19,-19) (28,21)};
          \addplot[blue!30, mark=text, text mark=afr] coordinates {(-6.51025020165559, 0.331557309537027)};
          \addplot[red!30, mark=text, text mark=aln] coordinates {(-0.155997995338138, -2.13522233336093)};
          \addplot[red!30, mark=text, text mark=arb] coordinates {(0.686887411262343, -15.6420716826835)};
          \addplot[red!30, mark=text, text mark=arz] coordinates {(6.06136495946954, -8.39855247999838)};
          \addplot[red!30, mark=text, text mark=ayr] coordinates {(19.7758570472401, 8.22555896713054)};
          \addplot[red!30, mark=text, text mark=ayr] coordinates {(7.90207042536995, 0.682459781242666)};
          \addplot[blue!30, mark=text, text mark=bba] coordinates {(-16.3741674946509, -15.2243206428482)};
          \addplot[blue!30, mark=text, text mark=ben] coordinates {(-4.21130498340626, -1.69836442516974)};
          \addplot[blue!30, mark=text, text mark=ben] coordinates {(-4.41603743254406, -0.363434900888657)};
          \addplot[red!30, mark=text, text mark=bqc] coordinates {(-15.0135917777278, -16.0135493174779)};
          \addplot[red!30, mark=text, text mark=bul] coordinates {(2.69177892460811, 1.19102873583881)};
          \addplot[red!30, mark=text, text mark=bul] coordinates {(-4.95101510564052, -6.60330187769513)};
          \addplot[blue!30, mark=text, text mark=cac] coordinates {(-4.60061022474516, 3.57572618577667)};
          \addplot[blue!30, mark=text, text mark=cak] coordinates {(4.19076047285395, 9.07042535075442)};
          \addplot[blue!30, mark=text, text mark=ceb] coordinates {(0.281369241719309, 7.01292183026532)};
          \addplot[blue!30, mark=text, text mark=ceb] coordinates {(3.36678572714387, 7.80076962300943)};
          \addplot[blue!30, mark=text, text mark=ceb] coordinates {(3.58550456585957, 6.85346966462827)};
          \addplot[red!30, mark=text, text mark=ces] coordinates {(7.88618914930384, -0.523924241668724)};
          \addplot[red!30, mark=text, text mark=ces] coordinates {(6.26171357372163, 4.58099192300774)};
          \addplot[red!30, mark=text, text mark=cmn] coordinates {(10.1438648223869, -17.6162296639417)};
          \addplot[blue!30, mark=text, text mark=cnh] coordinates {(-1.19109658386858, 3.98048615044214)};
          \addplot[blue!30, mark=text, text mark=cym] coordinates {(-3.22767331303148, -0.972522490490268)};
          \addplot[blue!30, mark=text, text mark=dan] coordinates {(-1.97880733438245, 0.376018818023161)};
          \addplot[black!90, mark=text, text mark=deu] coordinates {(-0.526667522339527, 8.57397368631263)};
          \addplot[black!90, mark=text, text mark=deu] coordinates {(-1.78651611229432, 4.46312331618974)};
          \addplot[black!90, mark=text, text mark=deu] coordinates {(-1.92839190398936, 2.96420016733532)};
          \addplot[black!90, mark=text, text mark=deu] coordinates {(9.07350322817029, 11.32083052444)};
          \addplot[black!90, mark=text, text mark=deu] coordinates {(7.80563819857427, 9.81702777626652)};
          \addplot[black!90, mark=text, text mark=deu] coordinates {(-4.87449757515021, -0.643459088500637)};
          \addplot[black!90, mark=text, text mark=deu] coordinates {(3.18437322680398, 3.95370200251333)};
          \addplot[black!90, mark=text, text mark=deu] coordinates {(6.87570747077113, 10.0956068911506)};
          \addplot[black!90, mark=text, text mark=deu] coordinates {(1.04708447182552, 5.16540015275616)};
          \addplot[black!90, mark=text, text mark=deu] coordinates {(8.83600671653109, 9.81653642033166)};
          \addplot[black!90, mark=text, text mark=deu] coordinates {(2.81209538188722, 3.06849646432269)};
          \addplot[blue!30, mark=text, text mark=ell] coordinates {(-1.63088698782592, -0.099490277988478)};
          \addplot[brown!90, mark=text, text mark=eng] coordinates {(-6.62378257420404, 0.767062467261383)};
          \addplot[brown!90, mark=text, text mark=eng] coordinates {(-10.1437777818839, -7.91425547894057)};
          \addplot[brown!90, mark=text, text mark=eng] coordinates {(-9.23301662114125, -1.41170182748587)};
          \addplot[brown!90, mark=text, text mark=eng] coordinates {(-7.00669278531551, -4.72446724804415)};
          \addplot[blue!30, mark=text, text mark=epo] coordinates {(-10.1061059074207, -8.16163271124925)};
          \addplot[red!30, mark=text, text mark=fin] coordinates {(0.878926354658096, -2.08455290949531)};
          \addplot[red!30, mark=text, text mark=fin] coordinates {(6.28142342069871, 1.73047698574235)};
          \addplot[red!30, mark=text, text mark=fin] coordinates {(8.71166357366405, -1.84141536606077)};
          \addplot[olive!90, mark=text, text mark=fra] coordinates {(-2.81187451428044, -1.86470301473557)};
          \addplot[olive!90, mark=text, text mark=fra] coordinates {(-0.255091817494266, -0.844374732644858)};
          \addplot[olive!90, mark=text, text mark=fra] coordinates {(-6.23292285333933, -5.2714083291672)};
          \addplot[olive!90, mark=text, text mark=fra] coordinates {(-1.90357326147768, -2.00423677787375)};
          \addplot[olive!90, mark=text, text mark=fra] coordinates {(0.788453173361469, 0.024735297954121)};
          \addplot[olive!90, mark=text, text mark=fra] coordinates {(-6.16280910366721, -1.49173284657312)};
          \addplot[olive!90, mark=text, text mark=fra] coordinates {(-7.87274321055869, -7.41208194181401)};
          \addplot[olive!90, mark=text, text mark=fra] coordinates {(-2.78430201238689, 0.358459082008912)};
          \addplot[olive!90, mark=text, text mark=fra] coordinates {(-11.8211985735506, -9.9771080241842)};
          \addplot[olive!90, mark=text, text mark=fra] coordinates {(3.21891638826211, 5.96925849345818)};
          \addplot[olive!90, mark=text, text mark=fra] coordinates {(1.68803890661228, 0.607538092971832)};
          \addplot[red!30, mark=text, text mark=guj] coordinates {(17.8917127645207, 16.0495914546579)};
          \addplot[blue!30, mark=text, text mark=gur] coordinates {(-2.08678262853449, 1.76233395570968)};
          \addplot[blue!30, mark=text, text mark=hat] coordinates {(-7.99988696769838, -4.33473740186292)};
          \addplot[blue!30, mark=text, text mark=hat] coordinates {(-6.89351753141293, -3.01109895948126)};
          \addplot[red!30, mark=text, text mark=hrv] coordinates {(9.41699983366105, -0.425393223882822)};
          \addplot[red!30, mark=text, text mark=hun] coordinates {(8.82596998459479, 0.204720644036477)};
          \addplot[red!30, mark=text, text mark=hun] coordinates {(12.4785318507343, 2.19879077665927)};
          \addplot[blue!30, mark=text, text mark=ind] coordinates {(-5.98031469114345, -0.39249039941385)};
          \addplot[blue!30, mark=text, text mark=ind] coordinates {(-5.48512001902267, -1.90588777973739)};
          \addplot[blue!30, mark=text, text mark=ita] coordinates {(-0.511759095290397, 0.214598729863336)};
          \addplot[blue!30, mark=text, text mark=ita] coordinates {(0.941540759213508, 3.41372923014834)};
          \addplot[blue!30, mark=text, text mark=ita] coordinates {(-0.40356292222444, 3.46877359263347)};
          \addplot[blue!30, mark=text, text mark=ita] coordinates {(3.56340160170969, 6.89882914920111)};
          \addplot[blue!30, mark=text, text mark=kek] coordinates {(-12.6247537172442, -2.92995871225523)};
          \addplot[blue!30, mark=text, text mark=kek] coordinates {(-11.4616286730321, -6.86233057610882)};
          \addplot[blue!30, mark=text, text mark=kjb] coordinates {(1.19931179235418, 6.26775844685589)};
          \addplot[red!30, mark=text, text mark=lat] coordinates {(2.32988528146413, -4.98866148636569)};
          \addplot[red!30, mark=text, text mark=lit] coordinates {(-3.29262326269535, -12.056036019176)};
          \addplot[blue!30, mark=text, text mark=mah] coordinates {(-14.0466116749839, -3.52008691727577)};
          \addplot[blue!30, mark=text, text mark=mam] coordinates {(-8.09056661768414, -1.99135967508009)};
          \addplot[blue!30, mark=text, text mark=mri] coordinates {(-7.81989274105555, -2.00056622892518)};
          \addplot[red!30, mark=text, text mark=mya] coordinates {(10.0019921499577, -0.036103586302061)};
          \addplot[blue!30, mark=text, text mark=nld] coordinates {(-1.82934856294165, 0.840576716750583)};
          \addplot[red!30, mark=text, text mark=nor] coordinates {(-7.00159381885381, -7.50628718789868)};
          \addplot[red!30, mark=text, text mark=nor] coordinates {(0.785240539249603, 0.491793532901408)};
          \addplot[red!30, mark=text, text mark=plt] coordinates {(0.473385246905299, -2.01967802559963)};
          \addplot[blue!30, mark=text, text mark=poh] coordinates {(0.625557868090787, 10.1412878994832)};
          \addplot[red!30, mark=text, text mark=por] coordinates {(-1.40815471560633, -2.18481771008159)};
          \addplot[blue!30, mark=text, text mark=por] coordinates {(-3.01481142110311, 0.086684961740285)};
          \addplot[blue!30, mark=text, text mark=por] coordinates {(-1.33570050343424, 2.86745669104564)};
          \addplot[red!30, mark=text, text mark=qub] coordinates {(6.32829411886459, -4.01560153541108)};
          \addplot[red!30, mark=text, text mark=quh] coordinates {(9.26985265716222, -2.37934681744998)};
          \addplot[red!30, mark=text, text mark=quy] coordinates {(8.94191659416049, 0.369098854469075)};
          \addplot[red!30, mark=text, text mark=quz] coordinates {(4.82657561556558, -6.3437739970615)};
          \addplot[blue!30, mark=text, text mark=ron] coordinates {(-3.24145180635513, -2.30717001272425)};
          \addplot[red!30, mark=text, text mark=rus] coordinates {(3.53662712301878, -3.36674865188475)};
          \addplot[blue!30, mark=text, text mark=som] coordinates {(3.70686120458128, 4.95483197148396)};
          \addplot[blue!30, mark=text, text mark=tbz] coordinates {(-14.5935323256255, -12.6941129764113)};
          \addplot[red!30, mark=text, text mark=tcw] coordinates {(25.2625692214628, 19.1162815329973)};
          \addplot[blue!30, mark=text, text mark=tgl] coordinates {(-4.12060539117675, 4.92564371926255)};
          \addplot[blue!30, mark=text, text mark=tlh] coordinates {(-11.7356929528241, -3.76115293195438)};
          \addplot[blue!30, mark=text, text mark=tpi] coordinates {(-12.2054859746898, -1.50643493772799)};
          \addplot[blue!30, mark=text, text mark=tpm] coordinates {(-12.8714721372639, -11.474434289126)};
          \addplot[red!30, mark=text, text mark=ukr] coordinates {(7.63388584435332, 2.62015177761905)};
          \addplot[red!30, mark=text, text mark=ukr] coordinates {(8.56067720996759, 2.77361774746852)};
          \addplot[blue!30, mark=text, text mark=vie] coordinates {(-1.14048287588808, 1.72035606851173)};
          \addplot[blue!30, mark=text, text mark=vie] coordinates {(-1.19171844098112, -0.319114946924168)};
          \addplot[blue!30, mark=text, text mark=vie] coordinates {(0.772894200362324, 3.5729217253914)};
          \addplot[red!30, mark=text, text mark=wal] coordinates {(-2.34491353848781, -5.89380241043079)};
          \addplot[blue!30, mark=text, text mark=wbm] coordinates {(-3.01079249601182, 4.614483660235)};
          \addplot[red!30, mark=text, text mark=xho] coordinates {(12.9204913342155, -0.570887600052081)};
          \addplot[red!30, mark=text, text mark=zom] coordinates {(-0.77346224058855, -1.23120316334402)};
      \end{axis}
    \end{tikzpicture}
  }
  \caption{Difficulties of 21 Europarl languages (left) and 106 Bibles (right), comparing difficulties when estimated from BPE-RNNLMs vs. char-RNNLMs. Highlighted on the right are deu and fra, for which we have many Bibles, and eng, which has often been prioritized even over these two in research. In the middle we see the difficulties of the 14 languages that are shared between the Bibles and Europarl aligned to each other (averaging all estimates), indicating that the general trends we see are not tied to either corpus.}
  \label{fig:language-difficulties-scatter}
\vspace*{-12pt}
\end{figure*}

The estimated difficulties are visualized in \cref{fig:language-difficulties-scatter}.
We can see that general trends are preserved between datasets: German and Hungarian are hardest, English and Lithuanian easiest.
As we can see in \cref{fig:ep-difficulties-modeltuning} for Europarl, the difficulty estimates are hardly affected when tuning the number of BPE merges per-language instead of globally, validating our approach of using the BPE model for our experiments.
A bigger difference seems to be the choice of char-RNNLM vs.\@ BPE-RNNLM, which changes the ranking of languages both on Europarl data and on Bibles. We still see German as the hardest language, but almost all other languages switch places. Specifically, we can see that the variance of the char-RNNLM is much higher.

\subsection{Are All Translations the Same?}

Texts like the Bible are justly infamous for their sometimes archaic or unrepresentative use of language.  The fact that we sometimes have multiple Bible translations in the same language lets us observe variation by translation style.

The sample standard deviation of $\langdifficulty_\idlang$ among the 106 Bibles $\idlang$ is 0.076/0.063 for BPE/char-RNNLM.  Within the 11 German, 11 French, and 4 English Bibles, the sample standard deviations were roughly 0.05/0.04, 0.05/0.04, and 0.02/0.04 respectively: so style accounts for less than half the variance.
We also consider another parallel corpus, created from the NIST OpenMT competitions on machine translation, in which each sentence has 4 English translations \citep{LDC2010T10,LDC2010T11,LDC2010T12,LDC2010T14,LDC2010T17,LDC2010T21,LDC2010T23,LDC2013T03,LDC2013T07}.
We get a sample standard deviation of 0.01/0.03 among the 4 resulting English corpora, suggesting that language difficulty estimates (particularly the BPE estimate) depend less on the translator,
to the extent that these corpora represent individual translators.

\section{What Correlates with Difficulty?}\label{sec:language-feature-regression}

Making use of our results on these languages, we can now answer the question: what features of a language correlate with the difference in language complexity?
Sadly, we cannot conduct all analyses on all data: the Europarl languages are well-served by existing tools like UDPipe \citep{straka_udpipe_2016}, but the languages of our Bibles are often not.
We therefore conduct analyses that rely on automatically extracted features only on the Europarl corpora. Note that to ensure a false discovery rate of at most $\alpha = .05$, all reported $p$-values have to be corrected using \citet{BenHoc95Controlling}'s procedure: only $p \le .05 \cdot \nicefrac{5}{28} \approx 0.009$ is significant.\looseness=-1

\paragraph{Morphological Counting Complexity}

\begin{figure}
  \centering
  \begin{adjustbox}{width=.475\linewidth}
    \begin{tikzpicture}
      \begin{axis}[
        xlabel={MCC},
        ylabel={difficulty ($\times 100$, BPE-RNNLM)},
        width=20em,
        height=23em,
        mark options={scale=0.8}]
        \addplot[black, mark=text, text mark=bg] coordinates {(96, -1.96707081398943)};
        \addplot[black, mark=text, text mark=cs] coordinates {(195, 2.03485880364099)};
        \addplot[black, mark=text, text mark=da] coordinates {(15, -4.17916721098119)};
        \addplot[black, mark=text, text mark=de] coordinates {(38, 4.50834112127343)};
        \addplot[black, mark=text, text mark=el] coordinates {(50, 1.95389189987913)};
        \addplot[black, mark=text, text mark=en] coordinates {(6, -1.03294277253079)};
        \addplot[black, mark=text, text mark=et] coordinates {(110, 0.834094638945209)};
        \addplot[black, mark=text, text mark=fi] coordinates {(198, -2.13595135068061)};
        \addplot[black, mark=text, text mark=fr] coordinates {(30, -0.513766100740676)};
        \addplot[black, mark=text, text mark=hu] coordinates {(94, 3.79841720282489)};
        \addplot[black, mark=text, text mark=it] coordinates {(52, 0.320577700153964)};
        \addplot[black, mark=text, text mark=lv] coordinates {(81, -3.51928809791452)};
        \addplot[black, mark=text, text mark=lt] coordinates {(152, -1.81850502795807)};
        \addplot[black, mark=text, text mark=nl] coordinates {(26, 1.75256772077613)};
        \addplot[black, mark=text, text mark=pl] coordinates {(112, 2.4908472217835)};
        \addplot[black, mark=text, text mark=pt] coordinates {(77, 1.38961398601107)};
        \addplot[black, mark=text, text mark=ro] coordinates {(60, 2.12949005572799)};
        \addplot[black, mark=text, text mark=sk] coordinates {(40, -1.64043426956102)};
        \addplot[black, mark=text, text mark=sl] coordinates {(100, -1.56520015850714)};
        \addplot[black, mark=text, text mark=es] coordinates {(71, 0.656470310773059)};
        \addplot[black, mark=text, text mark=sv] coordinates {(35, -4.12012127685736)};
      \end{axis}
    \end{tikzpicture}
  \end{adjustbox}
  \hfill
  \begin{adjustbox}{width=.475\linewidth}
    \begin{tikzpicture}
      \begin{axis}[
        xlabel={MCC},
        ylabel={difficulty ($\times 100$, char-RNNLM)},
        width=20em,
        height=23em,
        mark options={scale=0.8}]
        \addplot[black, mark=text, text mark=bg] coordinates {(96, -0.758781962981698)};
        \addplot[black, mark=text, text mark=cs] coordinates {(195, -0.264967920151023)};
        \addplot[black, mark=text, text mark=da] coordinates {(15, -2.15469603188326)};
        \addplot[black, mark=text, text mark=de] coordinates {(38, 10.7829209623774)};
        \addplot[black, mark=text, text mark=el] coordinates {(50, -1.40670782334514)};
        \addplot[black, mark=text, text mark=en] coordinates {(6, -4.77888806561201)};
        \addplot[black, mark=text, text mark=et] coordinates {(110, 2.83807085616185)};
        \addplot[black, mark=text, text mark=fi] coordinates {(198, -0.490584369689634)};
        \addplot[black, mark=text, text mark=fr] coordinates {(30, 4.44472942130247)};
        \addplot[black, mark=text, text mark=hu] coordinates {(94, 4.35054254568222)};
        \addplot[black, mark=text, text mark=it] coordinates {(52, -1.24817157280897)};
        \addplot[black, mark=text, text mark=lv] coordinates {(81, -7.30100221215144)};
        \addplot[black, mark=text, text mark=lt] coordinates {(152, -2.97984403754686)};
        \addplot[black, mark=text, text mark=nl] coordinates {(26, 1.20318609852111)};
        \addplot[black, mark=text, text mark=pl] coordinates {(112, 5.98798146340616)};
        \addplot[black, mark=text, text mark=pt] coordinates {(77, 1.59186227377252)};
        \addplot[black, mark=text, text mark=ro] coordinates {(60, -0.093545800443719)};
        \addplot[black, mark=text, text mark=sk] coordinates {(40, -4.44769414378188)};
        \addplot[black, mark=text, text mark=sl] coordinates {(100, -5.15316603734123)};
        \addplot[black, mark=text, text mark=es] coordinates {(71, 2.24595240298568)};
        \addplot[black, mark=text, text mark=sv] coordinates {(35, -4.21238841108621)};
      \end{axis}
    \end{tikzpicture}
  \end{adjustbox}
  \caption{MCC does not predict difficulty on Europarl. Spearman's $\rho$ is $.091$ / $.110$ with $p > .6$ for BPE-RNNLM (left) / char-RNNLM (right).}
  \label{fig:mcc-scatter}
\end{figure}
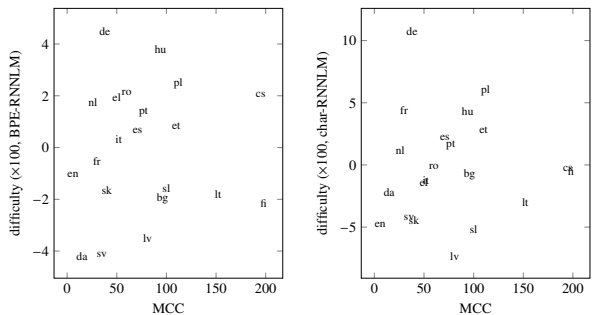

\citet{CotMieEis18All} suspected that inflectional morphology (i.e., the grammatical requirement to choose among forms like ``talk,'' ``talks,'' ``talking'') was mainly responsible for difficulty in modeling.  They found a language's Morphological Counting Complexity \citep{sagot2013comparing} to correlate positively with its difficulty.
We use the reported MCC values from that paper for our 21 Europarl languages, but to our surprise, find no statistically significant correlation with the newly estimated difficulties of our new language models.
Comparing the scatterplot for both languages in \cref{fig:mcc-scatter} with \citet{CotMieEis18All}'s Figure 1, we see that the high-MCC outlier Finnish has become much easier in our (presumably) better-tuned models.
We suspect that the reported correlation in that paper was mainly driven by such outliers and conclude that MCC is not a good predictor of modeling difficulty.  Perhaps finer measures of morphological complexity would be more predictive.

\paragraph{Head-POS Entropy}

\citet{dehouck-denis:2018:EMNLP} propose an alternative measure of morphosyntactic complexity. Given a corpus of dependency graphs, they estimate the conditional entropy of the POS tag of a random token's parent, conditioned on the\\token's type.  In a language where this \defn{HPE-mean} metric is low, most tokens can predict the POS of their parent even without context.  We compute HPE-mean from dependency parses of the Europarl data, generated using UDPipe 1.2.0 \citep{straka_udpipe_2016} \!and\! \mbox{freely-available\, tokenization,\, tagging,} parsing models trained on the Universal Dependencies 2.0 treebanks \citep{straka_tokenizing_2017}.%

HPE-mean may be regarded as the mean over all corpus tokens of \emph{Head POS Entropy} \cite{dehouck-denis:2018:EMNLP}, which is the entropy of the POS tag of a token's parent given that {\em particular} token's type.  We also compute \defn{HPE-skew}, the (positive) skewness of the empirical distribution of HPE on the corpus tokens.  We remark that in each language, HPE is 0 for most tokens.

As predictors of language difficulty, HPE-mean has a Spearman's $\rho = .004 / {-.045}$ ($p > .9 / .8$) and HPE-skew has a Spearman's $\rho = .032 / .158$ ($p > .8 / .4$), so this is not a positive result.

\paragraph{Average dependency length}

It has been observed that languages tend to minimize the distance between heads and dependents \citep{liu_dependency_2008}.
Speakers prefer shorter dependencies in both production and processing, and average dependency lengths tend to be much shorter than would be expected from randomly-generated parses \citep{futrell_large-scale_2015,liu_dependency_2017}.
On the other hand, there is substantial variability between languages, and it has been proposed, for example, that head-final languages and case-marking languages tend to have longer dependencies on average. \looseness=-1

Do language \emph{models} find short dependencies easier?
We find that average dependency lengths estimated from automated parses are very closely correlated with those estimated from (held-out) manual parse trees.
We again use the automatically-parsed Europarl data and compute dependency lengths using the \citet{futrell_large-scale_2015} procedure, which excludes punctuation and standardizes several other grammatical relationships (e.g., objects of prepositions are made to depend on their prepositions, and verbs to depend on their complementizers).
Our hypothesis that scrambling makes language harder to model seems confirmed at first: while the non-parametric (and thus more weakly powered) Spearman's $\rho = .196 / .092$ ($p = .394 / .691$), Pearson's $r  = .486 / .522$ ($p = .032 / .015$).  However, after correcting for multiple comparisons, this is also non-significant.\footnote{We also caution that the significance test for Pearson's assumes that the two variables are bivariate normal.  If not, then even a significant $r$ does not allow us to reject the null hypothesis of zero covariance \cite[Figs.\@ 1--2, \S5]{kowalski-1975}.}

\paragraph{WALS features}

The World Atlas of Language Structures \citep[WALS;][]{wals} contains nearly 200 binary and integer features for over 2000 languages.
Similarly to the Bible situation, not all features are present for all languages---and for some of our Bibles, no information can be found at all.
We therefore restrict our attention to two well-annotated WALS features that are present in enough of our Bible languages (foregoing Europarl to keep the analysis simple): 26A ``Prefixing vs. Suffixing in Inflectional Morphology'' and 81A ``Order of Subject, Object and Verb.''
The results are again not quite as striking as we would hope.  In particular, in Mood's median null hypothesis significance test neither 26A ($p > .3$ / $.7$ for BPE/char-RNNLM) nor 81A ($p > .6$ / $.2$ for BPE/char-RNNLM) show any significant differences between categories (detailed results in \cref{app:detailed-results-wals}).
We therefore turn our attention to much simpler, yet strikingly effective heuristics.\looseness=-1

\paragraph{Raw character sequence length}

An interesting correlation emerges between language difficulty for the char-RNNLM and the raw length in characters of the test corpus (detailed results in \cref{app:detailed-results-charseq}).
On both Europarl and the more reliable Bible corpus, we have positive correlation for the char-RNNLM at a significance level of $p<.001$, passing the multiple-test correction. The BPE-RNNLM correlation on the Bible corpus is very weak, suggesting that allowing larger units of prediction effectively eliminates this source of difficulty \cite{van2017multiscale}.

\paragraph{Raw word inventory}

Our most predictive feature, however, is the \emph{size of the word inventory}.
To obtain this number, we count the number of distinct types $|\mathcal{V}|$ in the (tokenized) training set of a language (detailed results in \cref{app:detailed-results-wordinventory}).%
\footnote{A more sophisticated version of this feature might consider not just the existence of certain forms but also their rates of appearance. We did calculate the entropy of the unigram distribution over words in a language, but we found that is strongly correlated with the size of the word inventory and not any more predictive.}
While again there is little power in the small set of Europarl languages, on the bigger set of Bibles we do see the biggest positive correlation of any of our features---but only on the BPE model ($p<1e$$-$$11$).
Recall that the char-RNNLM has no notion of words, whereas the number of BPE units increases with $|\mathcal{V}|$ (indeed, many whole words are BPE units, because we do many merges but BPE stops at word boundaries).  Thus, one interpretation is that the Bible corpora are too small to fit the parameters for all the units needed in large-vocabulary languages.  A similarly predictive feature on Bibles---whose numerator is this word inventory size---is the type/token ratio, where values closer to 1 are a traditional omen of undertraining.

An interesting observation is that on Europarl, the size of the word inventory and the morphological counting complexity of a language correlate quite well with each other (Pearson's $\rho = .693$ at $p = .0005$, Spearman's $\rho = .666$ at $p = .0009$), so the original claim in \citet{CotMieEis18All} about MCC may very well hold true after all.
Unfortunately, we cannot estimate the MCC for all the Bible languages, or this would be easy to check.\footnote{Perhaps in a future where more data has been annotated by the UniMorph project \citep{KIROV18}, a yet more comprehensive study can be performed, and the null hypothesis for the MCC can be ruled out after all.}

Given more nuanced linguistic measures (or more languages), our methods may permit discovery of specific linguistic correlates of modeling difficulty, beyond these simply suggestive results.

\section{Evaluating Translationese}\label{sec:translationese-results}

Our previous experiments treat translated sentences just like natively generated sentences.  But since Europarl contains information about which language an intent was originally expressed in,\footnote{It should be said that using Europarl for translationese studies is not without caveats \citep{rabinovich2016parallel}, one of them being the fact that not all language pairs are translated equally: a natively Finnish sentence is translated first into English, French, or German (\defn{pivoting}) and only from there into any other language like Bulgarian.} here we have the opportunity to ask another question: is translationese harder, easier, indistinguishable, or impossible to tell? We tackle this question by splitting each language $\idlang$ into two sub-languages, ``native'' $\idlang$ and ``translated'' $\idlang$, resulting in 42 sub-languages with 42 difficulties.\footnote{This method would also allow us to study the effect of source language, yielding
  $\langdifficulty_{\idlang \leftarrow \idlang'}$ for sentences translated from $\idlang'$ into $\idlang$.
  Similarly, we could have included surprisals from \emph{both} models,
  \emph{jointly} estimating $\langdifficulty_{\idlang,\text{char-RNN}}$ and $\langdifficulty_{\idlang,\text{BPE}}$ values.}
Each intent is expressed in at most 21 sub-languages, so this approach \emph{requires} a regresssion method that can handle missing data, such as the probabilistic approach we proposed in \cref{sec:aggregation-regression}.
Our mixed-effects modeling ensures that our estimation focuses on the differences between languages, controlling for content by automatically fitting the $n_i$ factors.
Thus, we are not in danger of calling native German more complicated than translated German just because German speakers in Parliament may like to talk about complicated things in complicated ways.

In a first attempt, we simply use our already-trained BPE-best models (as they perform the best and are thus most likely to support claims about the language itself rather than the shortcomings of any singular model), limit ourselves to only splitting the eight languages that have at least 500 native sentences\footnote{en (3256), fr (1650), de (1275), pt (1077), it (892), es (685), ro (661), pl (594)} (to ensure stable results).  Indeed we \emph{seem} to find that native sentences are slightly more difficult: their $\langdifficulty_\idlang$ is 0.027 larger ($\pm$ 0.023, averaged over our selected 8 languages).  

But are they? This result is confounded by the fact that our RNN language models \emph{were trained} mostly on translationese text (even the English data is mostly translationese).
Thus, translationese might merely be \emph{different} \citep{TACL618}---not necessarily easier to model, but overrepresented when training the model, making the translationese test sentences more predictable.
To remove this confound, we must train our language models on equal parts translationese and native text.  We cannot do this for multiple languages
at once, given our requirement of training all language models on the same intents.
We thus choose to balance only \emph{one} language---we train all models for all languages, making sure that the training set for one language is balanced---and then perform our regression, reporting the translationese and native difficulties only for the balanced language.
We repeat this process for every language that has enough intents.
We sample equal numbers of native and non-native sentences, such that there are $\sim$1M words in the corresponding English column (to be comparable to the PTB size).  To raise the number of languages we can split in this way, we restrict ourselves here to fully-parallel Europarl in only 10 languages\footnote{da, de, en, es, fi, fr, it, nl, pt, sv} instead of 21, thus ensuring that each of these 10 languages has enough native sentences.
\looseness=-1

On this level playing field, the previously observed effect practically disappears (-0.0044 $\pm$ 0.022), leading us to question the widespread hypothesis that translationese is ``easier'' to model \citep{baker1993corpus}.
\footnote{Of course we \emph{cannot} claim that it is just as hard to \emph{read} or \emph{translate} as native text---those are different claims altogether---but only that it is as easy to monolingually language-model.}

\section{Conclusion}\label{sec:conclusion}

\vspace{-1pt}
There is a real danger in cross-linguistic studies of over-extrapolating from limited data.  We re-evaluated the conclusions of  \citet{CotMieEis18All} on a larger set of languages, requiring new methods to select fully parallel data (\cref{sec:bible-selection}) or handle missing data.  We showed how to fit a paired-sample multiplicative mixed-effects model to probabilistically obtain language difficulties from at-least-pairwise parallel corpora.
Our language difficulty estimates were largely stable across datasets and language model architectures, but they were not significantly predicted by linguistic factors.
However, a language's vocabulary size and the length in characters of its sentences were well-correlated with difficulty on our large set of languages.  Our mixed-effects approach could be used to assess other NLP systems via parallel texts, separating out the influences on performance of language, sentence, model architecture, and training procedure.

\section*{Acknowledgments}

\vspace{-1pt}
This work was supported by the National Science Foundation under Grant No.\@ 1718846.

% Do not number the acknowledgment section. Do not include this section when submitting your paper for review.

\bibliography{multilm}
\bibliographystyle{acl_natbib}

\newpage
\appendix

\section{A Note on Missing Data}\label{app:missingness}
We stated that our model can deal with missing data, but this is true only for the case of data \defn{missing completely at random} (MCAR), the strongest assumption we can make about missing data: the missingness of data is neither influenced by what the value would have been (had it not been missing), nor by any covariates.
Sadly, this assumption is rarely met in real translations, where difficult, useless, or otherwise \emph{distinctive} sentences may be skipped.  This leads to data \defn{missing at random} (MAR), where the missingness of a translation is correlated with the original sentence it should have been translated from---or even data \defn{missing not at random} (MNAR), where the missingness of a translation is correlated with that translation, i.e., the original sentence was translated, but the translation was then deleted for a reason that depends on the translation itself).
For this reason we use fully parallel data where possible; in fact, we only make use of the ability to deal with missing data in \cref{sec:translationese-results}.%
\footnote{Note that this application counts as data MAR and not MCAR, thus technically violating our requirements, but only in a minor enough way that we are confident it can still be applied.}

\section{Regression, Model 3: Handling outliers cleverly}\label{app:model-3}

Consider the problem of outliers.  In some cases, sloppy translation will yield a $y_\idlangsent$ that is unusually high or low given the $y_\idlangsent'$ values of other languages $\idlang'$.  Such a $y_\idlangsent$ is not good evidence of the quality of the language model for language $\idlang$ since it has been corrupted by the sloppy translation.  However, under Model 1 or 2, we could not simply explain this corrupted $y_\idlangsent$ with the random residual $\epsilon_\idlangsent$ since large $|\epsilon_\idlangsent|$ is highly unlikely under the Gaussian assumption of those models.  Rather, $y_\idlangsent$ would have significant influence on our estimate of the per-language effect $\langdifficulty_\idlang$.
This is the usual motivation for switching to L1 regression, which replaces the Gaussian prior on the residuals with a Laplace prior.\footnote{An alternative would be to use a method like RANSAC to discard $y_\idlangsent$ values that do not appear to fit.}

How can we include this idea into our models?
First let us identify two failure modes:
\begin{enumerate}[(a)]
  \item part of a sentence was omitted (or added) during translation, changing the $\sentunits_\idsent$ additively; thus we should use a noisy $\sentunits_\idsent + \nu_\idlangsent$ in place of $\sentunits_\idsent$ in \cref{eqn:y_n_d_e,eqn:sigma_i}
  \item the style of the translation was unusual throughout the sentence; thus we should use a noisy $\sentunits_\idsent \cdot \exp \nu_\idlangsent$ instead of $\sentunits_\idsent$ in \cref{eqn:y_n_d_e,eqn:sigma_i}
\end{enumerate}
In both cases $\nu_\idlangsent \sim \mathrm{Laplace}(0, b)$, i.e., $\nu_\idlangsent$ specifies sparse additive or multiplicative noise in $\nu_\idlangsent$ (on language $j$ only).\footnote{However, version (a) is then deficient since it then incorrectly allocates some probability mass to $\sentunits_\idsent + \nu_\idlangsent < 0$ and thus $y_\idlangsent < 0$ is possible.  This could be fixed by using a different sparsity-inducing distribution.}

Let us write out version (b), which is a modification of Model 2 (\cref{eqn:y_n_d_e,eqn:sigma_i,eqn:fw}):
\begin{align}
  y_\idlangsent &= (\sentunits_\idsent \cdot \exp \nu_\idlangsent) \cdot \exp(\langdifficulty_\idlang) \cdot \exp (\epsilon_\idlangsent)\nonumber\\
  &= \sentunits_\idsent \cdot \exp(\langdifficulty_\idlang) \cdot \exp (\epsilon_\idlangsent + \nu_\idlangsent)\label{eqn:y_n_d_e_nu}\\
  \nu_\idlangsent &\sim \mathrm{Laplace}(0, b)\\
  \sigma_\idsent^2 &= \ln \Big( 1 + \tfrac{\exp(\sigma^2) - 1}{\sentunits_\idsent \cdot \exp \nu_\idlangsent} \Big)\\
  \epsilon_\idlangsent &\sim \mathcal{N}\Big( \tfrac{\sigma^2 - \sigma_\idsent^2}{2}\,,\; \sigma_\idsent^2 \Big),
\end{align}
Comparing \cref{eqn:y_n_d_e_nu} to \cref{eqn:y_n_d_e}, we see that we are now modeling the residual error in $\log y_\idlangsent$ as a \emph{sum of two noise terms} $a_\idlangsent = \nu_\idlangsent + \epsilon_\idlangsent$ and penalizing it by (some multiple of) the weighted sum of $|\nu_\idlangsent|$ and $\epsilon_\idlangsent^2$, where large errors can be more cheaply explained using the former summand, and small errors using the latter summand.%
\footnote{%
  The cheapest penalty or explanation of the weighted sum $\delta|\nu_\idlangsent| + \frac{1}{2}\epsilon_\idlangsent^2$ for some weighting or threshold $\delta$ (which adjusts the relative variances of the two priors) is $\nu = 0$ if $|a| \leq \delta$, $\nu = a - \delta$ if $a \geq \delta$, and $\nu = -(a-\delta)$ if $a < -\delta$ (found by minimizing $\delta|\nu| + \frac{1}{2}(a - \nu)^2$, a convex function of $\nu$).
  This implies that we incur a quadratic penalty $\frac{1}{2}a^2$ if $|a| \leq \delta$, and a linear penalty $\delta(|a| - \frac{1}{2}\delta)$ for the other cases; this penalty function is exactly the Huber loss of $a$, and essentially imposes an L2 penalty on small residuals and an L1 penalty on large residuals (outliers), so our estimate of $\langdifficulty_\idlang$ will be something between a mean and a median.
}
The weighting of the two terms is a tunable hyperparameter.

We did implement this model and test it on data, but not only was fitting it much harder and slower, it also did not yield particularly encouraging results, leading us to omit it from the main text.

\section{Goodness of fit of our difficulty estimation models}\label{app:goodness-of-fit-plot}

\begin{figure*}
  \includegraphics[width=\linewidth]{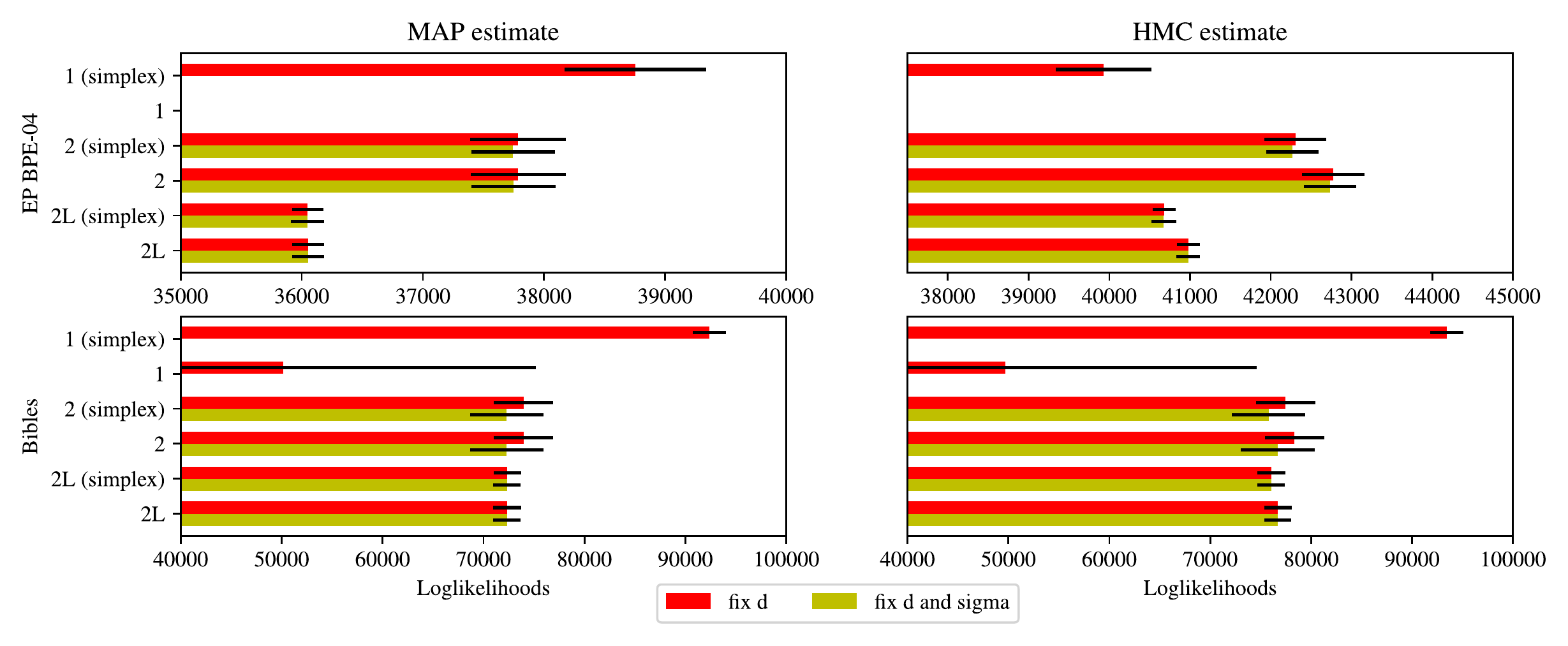}
  \caption{Achieved log-likelihoods on held-out data. Top: Europarl (BPE), Bottom: Bibles, Left: MAP inference, Right: HMC inference (posterior mean).}
  \label{fig:goodness-of-fit}
  \vspace{-\baselineskip}
\end{figure*}

\Cref{fig:goodness-of-fit} shows the log-probability of held-out data under the regression model, by fixing the estimated difficulties $\langdifficulty_\idlang$ (and sometimes also the estimated variance $\sigma^2$) to their values obtained from training data, and then finding either MAP estimates or posterior means (by running HMC using STAN) of the other parameters, in particular $\sentunits_\idsent$ for the new sentences $i$. The error bars are the standard deviations when running the model over different subsets of data.
The ``simplex'' versions of regression in \cref{fig:goodness-of-fit} force all $\langdifficulty_\idlang$ to add up to the number of languages (i.e., encouraging each one to stay close to 1). This is \emph{necessary} for Model 1, which otherwise is unidentifiable (hence the enormous standard deviation).  For other models, it turns out to only have much of an effect on the posterior means, not on the log-probability of held out data under the MAP estimate.
For stability, we in all cases take the best result when initializing the new parameters randomly or ``sensibly,'' i.e., the $\sentunits_\idsent$ of an intent $\idsent$ is initialized as the average of the corresponding sentences' $y_\idlangsent$.

\section{Data selection: Europarl}\label{app:europarl-extraction}

In the ``Corrected \& Structured Europarl Corpus'' (CoStEP) corpus \citep{GraBatVol14Cleaning}, sessions are grouped into \textit{turns}, each turn has one speaker (that is marked with clean attributes like native language) and a number of aligned \emph{paragraphs} for each language, i.e., the actual multitext.

\begin{figure}
  \begin{adjustbox}{width=\linewidth}
    \begin{tikzpicture}
      \begin{axis}[
          axis x line*=bottom, axis y line*=left,
          ybar=3pt,
          ymin=0,
          ylabel={\% of intents},
          y label style={at={(axis description cs:-0.06,0.5)}},
          xlabel={\#languages parallel},
          bar width=10pt,
          ymajorgrids,
          yminorgrids,
          enlarge x limits={value=0.025, auto},
          xtick=data,
          xticklabel style={text height=.7em},
          nodes near coords align={horizontal}, every node near coord/.append style={rotate=90},
          width=30em, height=15em,
          ]
        \addplot[red,fill=orange] coordinates {(1,0.01) (2,0.01) (3,0.47) (4,0.33) (5,0.48) (6,0.16) (7,0.21) (8,0.71) (9,4.56) (10,15.56) (11,25.64) (12,0.02) (13,0.00) (14,0.33) (15,0.01) (16,0.33) (17,0.61) (18,3.52) (19,4.73) (20,1.79) (21,13.73)};
      \end{axis}
    \end{tikzpicture}
  \end{adjustbox}
  \caption{In how many languages are the intents in Europarl translated? (intents from ill-fitting turns included in 100\%, but not plotted)}
  \label{fig:ep:multi-ness}
\end{figure}
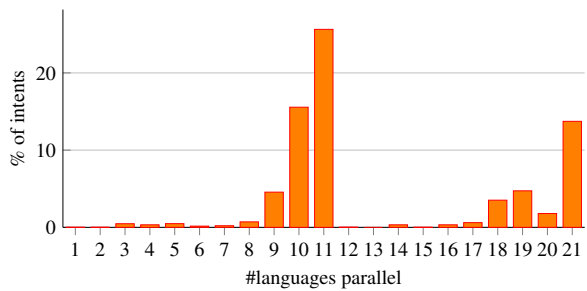

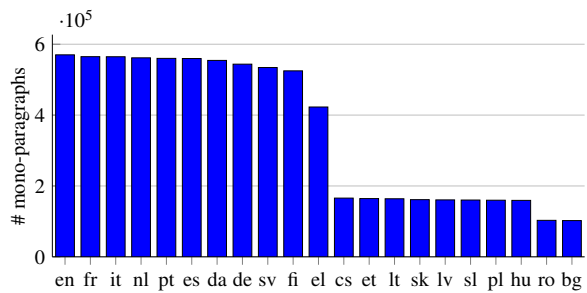
\begin{figure}
  \begin{adjustbox}{width=\linewidth}
    \begin{tikzpicture}
      \begin{axis}[
          axis x line*=bottom, axis y line*=left,
          ybar=3pt,
          ymin=0,
          ylabel={\# mono-paragraphs},
          y label style={at={(axis description cs:-0.03,0.5)}},
          bar width=10pt,
          ymajorgrids,
          yminorgrids,
          symbolic x coords={en,fr,it,nl,pt,es,da,de,sv,fi,el,cs,et,lt,sk,lv,sl,pl,hu,ro,bg},
          enlarge x limits={value=0.025, auto},
          xtick=data,
          xticklabel style={text height=.7em},
          nodes near coords align={horizontal}, every node near coord/.append style={rotate=90},
          width=30em, height=15em
          ]
        \addplot[black,fill=blue] coordinates {(en,570072) (fr,564905) (it,564704) (nl,561631) (pt,560077) (es,559644) (da,554342) (de,543766) (sv,534246) (fi,524872) (el,422868) (cs,165740) (et,164374) (lt,163852) (sk,161355) (lv,160678) (sl,160276) (pl,159778) (hu,159325) (ro,102965) (bg,102274)};
      \end{axis}
    \end{tikzpicture}
  \end{adjustbox}
  \caption{How many sentences are there per Europarl language?}
  \label{fig:ep:mono-langs}
\end{figure}

We ignore all paragraphs that are in \emph{ill-fitting} turns (i.e., turns with an unequal number of paragraphs across languages, a clear sign of an incorrect alignment), losing roughly 27\% of intents.
After this cleaning step, only 14\% of \emph{intents} are represented in all 21 languages, see the distribution in \cref{fig:ep:multi-ness} (the peak at 11 languages is explained by looking at the raw number of sentences present in each language, shown in \cref{fig:ep:mono-langs}).

\begin{figure}
  \begin{adjustbox}{width=\linewidth}
    \begin{tikzpicture}
      \begin{axis}[
          axis x line*=bottom, axis y line*=left,
          ybar=3pt,
          ymin=0,
          ylabel={\% native of sentences},
          y label style={at={(axis description cs:-0.05,0.5)}},
          xlabel={languages, sorted by absolute \# native sentences},
          bar width=10pt,
          ymajorgrids,
          yminorgrids,
          symbolic x coords={en,fr,de,es,nl,it,pt,sv,el,fi,pl,da,ro,hu,sk,cs,sl,lt,bg,et,lv},
          enlarge x limits={value=0.025, auto},
          xtick=data,
          xticklabel style={text height=.7em},
          nodes near coords align={horizontal}, every node near coord/.append style={rotate=90},
          width=30em, height=15em
          ]
        \addplot[olive,fill=green] coordinates {(en,19.32) (fr,12.67) (de,11.62) (es,7.26) (nl,5.61) (it,5.45) (pt,5.14) (sv,3.50) (el,2.99) (fi,1.77) (pl,5.39) (da,1.06) (ro,5.14) (hu,1.87) (sk,1.67) (cs,1.23) (sl,1.14) (lt,1.11) (bg,0.85) (et,0.40) (lv,0.18)};
      \end{axis}
    \end{tikzpicture}
  \end{adjustbox}
  \caption{How many of the Europarl sentences in one language are ``native''?}
  \label{fig:ep:nativity}
\end{figure}
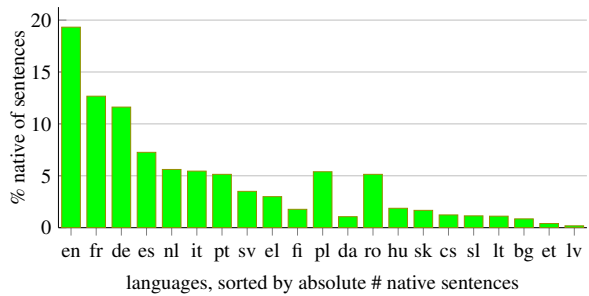

Since we want a fair comparison, we use the aforementioned 14\% of Europarl, giving us 78169 intents that are represented in all 21 languages.

Finally, it should be said that the text in CoStEP itself contains some markup, marking reports, ellipses, etc., but we strip this additional markup to obtain the raw text.
We tokenize it using the reversible language-agnostic tokenizer of \citet{MieEis18Spell}\footnote{\url{http://sjmielke.com/papers/tokenize/}} and split the obtained 78169 paragraphs into training set, development set for tuning our language models, and test set for our regression, again by dividing the data into blocks of 30 paragraphs and then taking 5 sentences for the development and test set each, leaving the remainder for the training set. This way we ensure uniform division over sessions of the parliament and sizes of $\nicefrac{2}{3}$, $\nicefrac{1}{6}$, and $\nicefrac{1}{6}$, respectively.

\subsection{\hspace{-6pt}How are the source languages distributed?}

An obvious question we should ask is: how many ``native'' sentences can we actually find in Europarl?
One could assume that there are as many native \emph{sentences} as there are \emph{intents} in total, but there are three issues with this:
the first is that the \emph{president} in any Europarl session is never annotated with name or native language (leaving us guessing what the native version of any president-uttered intent is; 12\% of all intents in Europarl that can be extracted have this problem), the second is that a number of speakers are labeled with ``unknown'' as native language (10\% of sentences), and finally some speakers have their native language annotated, but it is nowhere to be found in the corresponding sentences (7\% of sentences).

Looking only at the native sentences that we could identify, we can see that there are native sentences in every language, but unsurprisingly, some languages are overrepresented. Dividing the number of \emph{native} sentences in a language by the number of \emph{total} sentences, we get an idea of how ``natively spoken'' the language is in Europarl, shown in \cref{fig:ep:nativity}.

\section{Data selection: Bibles}\label{app:bible-selection}

The Bible is composed of the \emph{Old Testament} and the \emph{New Testament} (the latter of which has been much more widely translated), both consisting of individual \emph{books}, which, in turn, can be separated into \emph{chapters}, but we will only work with the smallest subdivision unit: the \emph{verse}, corresponding roughly to a sentence.
Turning to the collection assembled by \citet{MayCys14Creating}, we see that it has over 1000 New Testaments, but far fewer complete Bibles.

Despite being a fairly standardized book, not all Bibles are fully parallel.
Some verses and sometimes entire books are missing in some Bibles---some of these discrepancies may be reduced to the question of the legitimacy of certain biblical books, others are simply artifacts of verse numbering and labeling of individual translations.

\begin{figure}
  \centering
  \begin{minipage}{.45\linewidth}
    \centering
    \includegraphics[height=15em,interpolate=false]{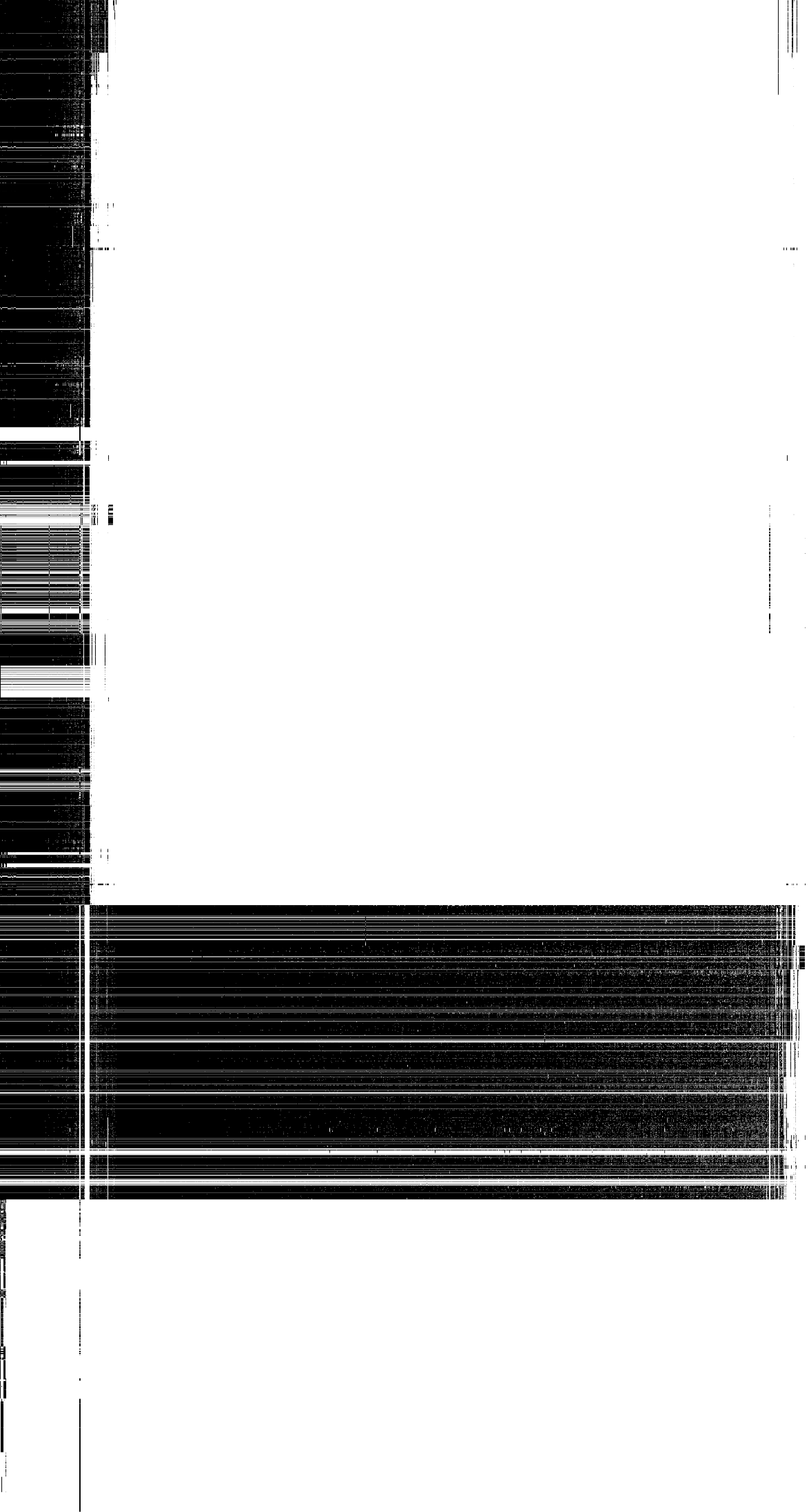}
    \subcaption{All 1174 Bibles, in packets of 20 verses, Bibles sorted by number of verses present, verses in chronological order. The New Testament (third quarter of verses) is present in almost every Bible.}
  \end{minipage}
  \hspace*{1em}
  \begin{minipage}{.46\linewidth}
    \centering
    \includegraphics[height=15em,interpolate=false]{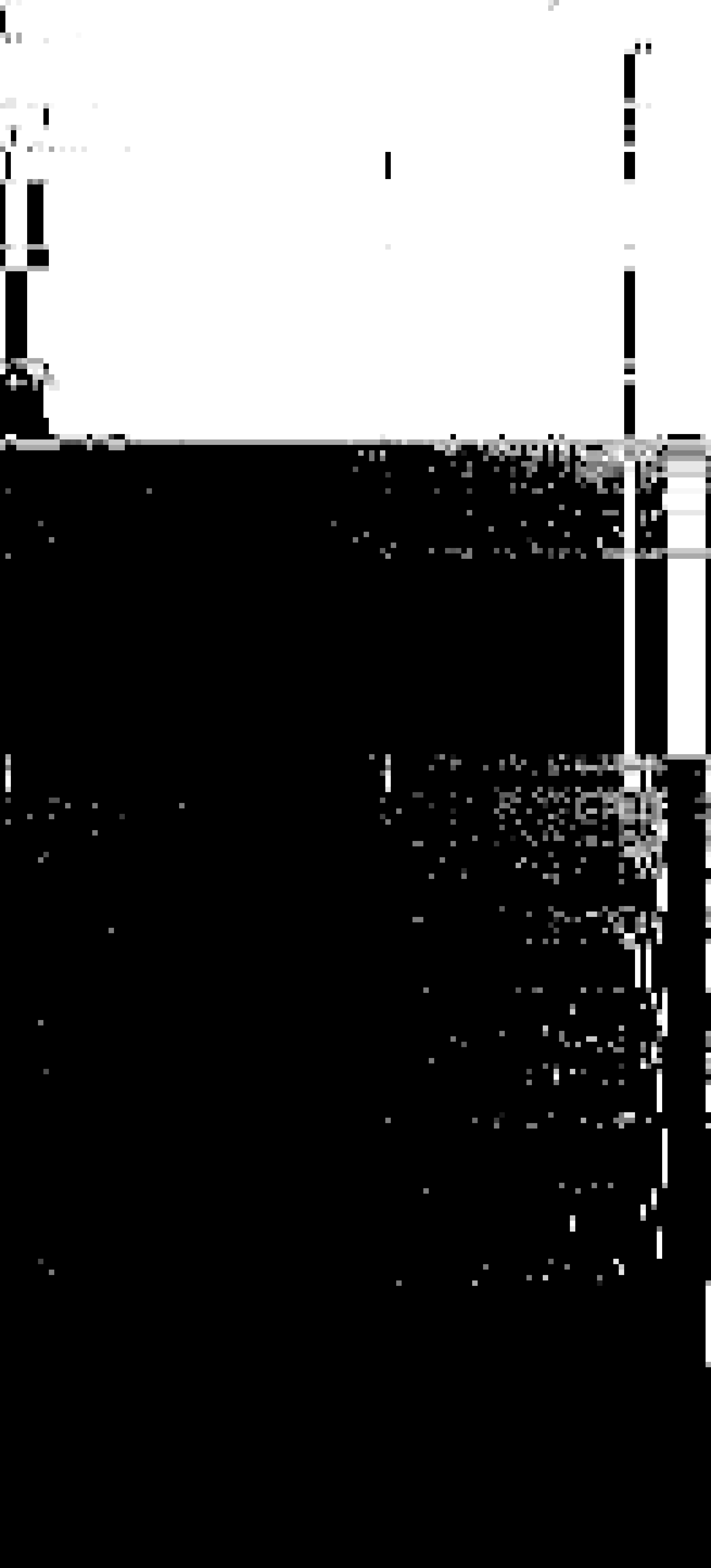}
    \subcaption{The 131 Bibles with at least 20000 verses, in packets of 150 verses (this time, both sorted). The optimization task is to remove rows and columns in this picture until only black remains.}
  \end{minipage}
  \caption{Presence (black) of verses (y-axis) in Bibles (x-axis). Both pictures are downsampled, resulting in grayscale values for all packets of N values.}
  \label{fig:bible-matrix}
  \vspace{-\baselineskip}
\end{figure}

For us, this means that we can neither simply take all translations that have ``the entire thing'' (in fact, no single Bible in the set covers the union of all others' verses), nor can we take all Bibles and work with the verses that they all share (because, again, no single verse is shared over all given Bibles).
The whole situation is visualized in \cref{fig:bible-matrix}.

We have to find a tradeoff: take as many Bibles as possible that share as many verses as possible.
Specifically, we cast this selection process as an optimization problem: select Bibles such that the number of verses overall (i.e., the number of verses shared times the number of Bibles) is maximal, breaking ties in favor of including more Bibles and ensuring that we have at least 20000 verses overall to ensure applicability of neural language models.
This problem can be cast as an integer linear program and solved using a standard optimization tool (Gurobi) within a few hours.

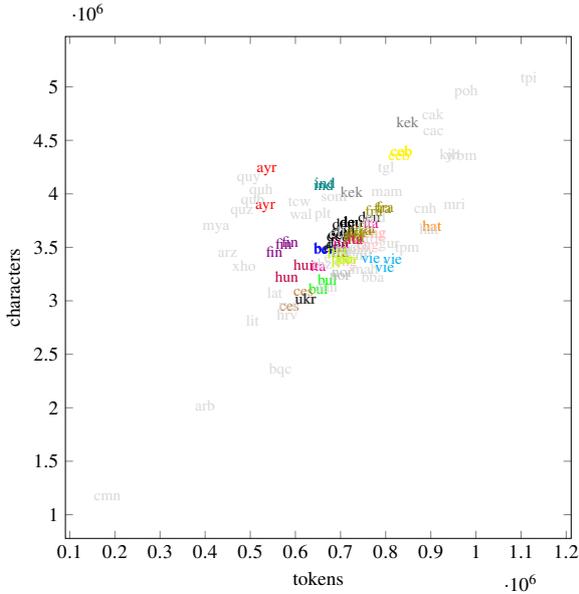
\begin{figure}
  \begin{adjustbox}{width=\linewidth}
    \begin{tikzpicture}
      \begin{axis}[
          xlabel={tokens},
          ylabel={characters},
          width=30em,
          height=30em,
          legend style={at={(0.5, -.1)},anchor=north,draw=none,fill=none,column sep=.1em,name=legend},
          legend columns=7,
          legend cell align=left,
          mark options={scale=0.8}]
          \addplot[smooth,gray!30,mark=text,text mark=afr] coordinates {(767658, 3579519)};
          \addplot[smooth,gray!30,mark=text,text mark=aln] coordinates {(741734, 3503572)};
          \addplot[smooth,gray!30,mark=text,text mark=arb] coordinates {(400018, 2026941)};
          \addplot[smooth,gray!30,mark=text,text mark=arz] coordinates {(449685, 3439440)};
          \addplot[smooth,red,mark=text,text mark=ayr] coordinates {(536709, 4217454)};
          \addplot[smooth,red,mark=text,text mark=ayr] coordinates {(533301, 3878169)};
          \addplot[smooth,gray!30,mark=text,text mark=bba] coordinates {(771586, 3223007)};
          \addplot[smooth,blue,mark=text,text mark=ben] coordinates {(666934, 3493541)};
          \addplot[smooth,blue,mark=text,text mark=ben] coordinates {(665635, 3491451)};
          \addplot[smooth,gray!30,mark=text,text mark=bqc] coordinates {(566769, 2357033)};
          \addplot[smooth,green,mark=text,text mark=bul] coordinates {(670939, 3198939)};
          \addplot[smooth,green,mark=text,text mark=bul] coordinates {(650848, 3117419)};
          \addplot[smooth,gray!30,mark=text,text mark=cac] coordinates {(906168, 4580436)};
          \addplot[smooth,gray!30,mark=text,text mark=cak] coordinates {(904854, 4750215)};
          \addplot[smooth,yellow,mark=text,text mark=ceb] coordinates {(830003, 4362620)};
          \addplot[smooth,yellow,mark=text,text mark=ceb] coordinates {(834874, 4397327)};
          \addplot[smooth,yellow,mark=text,text mark=ceb] coordinates {(835244, 4399006)};
          \addplot[smooth,brown,mark=text,text mark=ces] coordinates {(586832, 2938519)};
          \addplot[smooth,brown,mark=text,text mark=ces] coordinates {(618091, 3083220)};
          \addplot[smooth,gray!30,mark=text,text mark=cmn] coordinates {(183730, 1169488)};
          \addplot[smooth,gray!30,mark=text,text mark=cnh] coordinates {(888337, 3871777)};
          \addplot[smooth,gray!30,mark=text,text mark=cym] coordinates {(730105, 3404119)};
          \addplot[smooth,gray!30,mark=text,text mark=dan] coordinates {(668435, 3134527)};
          \addplot[smooth,black,mark=text,text mark=deu] coordinates {(723790, 3736702)};
          \addplot[smooth,black,mark=text,text mark=deu] coordinates {(724431, 3743085)};
          \addplot[smooth,black,mark=text,text mark=deu] coordinates {(723809, 3735577)};
          \addplot[smooth,black,mark=text,text mark=deu] coordinates {(688011, 3510094)};
          \addplot[smooth,black,mark=text,text mark=deu] coordinates {(764178, 3781573)};
          \addplot[smooth,black,mark=text,text mark=deu] coordinates {(693459, 3530995)};
          \addplot[smooth,black,mark=text,text mark=deu] coordinates {(697472, 3629224)};
          \addplot[smooth,black,mark=text,text mark=deu] coordinates {(694744, 3609445)};
          \addplot[smooth,black,mark=text,text mark=deu] coordinates {(707832, 3658622)};
          \addplot[smooth,black,mark=text,text mark=deu] coordinates {(707029, 3720910)};
          \addplot[smooth,black,mark=text,text mark=deu] coordinates {(707676, 3650163)};
          \addplot[smooth,gray!30,mark=text,text mark=ell] coordinates {(693645, 3648631)};
          \addplot[smooth,pink,mark=text,text mark=eng] coordinates {(776170, 3603294)};
          \addplot[smooth,pink,mark=text,text mark=eng] coordinates {(775736, 3615463)};
          \addplot[smooth,pink,mark=text,text mark=eng] coordinates {(760378, 3499676)};
          \addplot[smooth,pink,mark=text,text mark=eng] coordinates {(712354, 3342818)};
          \addplot[smooth,gray!30,mark=text,text mark=epo] coordinates {(689205, 3377765)};
          \addplot[smooth,violet,mark=text,text mark=fin] coordinates {(588091, 3554233)};
          \addplot[smooth,violet,mark=text,text mark=fin] coordinates {(574382, 3531997)};
          \addplot[smooth,violet,mark=text,text mark=fin] coordinates {(553226, 3460382)};
          \addplot[smooth,olive,mark=text,text mark=fra] coordinates {(730505, 3613692)};
          \addplot[smooth,olive,mark=text,text mark=fra] coordinates {(733739, 3601814)};
          \addplot[smooth,olive,mark=text,text mark=fra] coordinates {(757891, 3684949)};
          \addplot[smooth,olive,mark=text,text mark=fra] coordinates {(798778, 3880462)};
          \addplot[smooth,olive,mark=text,text mark=fra] coordinates {(699376, 3460400)};
          \addplot[smooth,olive,mark=text,text mark=fra] coordinates {(750314, 3667818)};
          \addplot[smooth,olive,mark=text,text mark=fra] coordinates {(730266, 3604294)};
          \addplot[smooth,olive,mark=text,text mark=fra] coordinates {(773832, 3840763)};
          \addplot[smooth,olive,mark=text,text mark=fra] coordinates {(793373, 3872449)};
          \addplot[smooth,olive,mark=text,text mark=fra] coordinates {(736166, 3649028)};
          \addplot[smooth,olive,mark=text,text mark=fra] coordinates {(723979, 3578768)};
          \addplot[smooth,gray!30,mark=text,text mark=guj] coordinates {(701115, 3484213)};
          \addplot[smooth,gray!30,mark=text,text mark=gur] coordinates {(807912, 3514728)};
          \addplot[smooth,orange,mark=text,text mark=hat] coordinates {(902953, 3705206)};
          \addplot[smooth,gray!30,mark=text,text mark=hat] coordinates {(896368, 3675548)};
          \addplot[smooth,gray!30,mark=text,text mark=hrv] coordinates {(582123, 2872892)};
          \addplot[smooth,purple,mark=text,text mark=hun] coordinates {(580857, 3221829)};
          \addplot[smooth,purple,mark=text,text mark=hun] coordinates {(621960, 3341969)};
          \addplot[smooth,teal,mark=text,text mark=ind] coordinates {(665851, 4100246)};
          \addplot[smooth,teal,mark=text,text mark=ind] coordinates {(662453, 4084737)};
          \addplot[smooth,magenta,mark=text,text mark=ita] coordinates {(650711, 3326932)};
          \addplot[smooth,magenta,mark=text,text mark=ita] coordinates {(767342, 3725108)};
          \addplot[smooth,magenta,mark=text,text mark=ita] coordinates {(701075, 3540143)};
          \addplot[smooth,magenta,mark=text,text mark=ita] coordinates {(732168, 3579433)};
          \addplot[smooth,gray,mark=text,text mark=kek] coordinates {(848512, 4671774)};
          \addplot[smooth,gray,mark=text,text mark=kek] coordinates {(725055, 4017448)};
          \addplot[smooth,gray!30,mark=text,text mark=kjb] coordinates {(941765, 4362269)};
          \addplot[smooth,gray!30,mark=text,text mark=lat] coordinates {(554470, 3078615)};
          \addplot[smooth,gray!30,mark=text,text mark=lit] coordinates {(504906, 2814757)};
          \addplot[smooth,gray!30,mark=text,text mark=mah] coordinates {(754407, 3302396)};
          \addplot[smooth,gray!30,mark=text,text mark=mam] coordinates {(803027, 4014871)};
          \addplot[smooth,gray!30,mark=text,text mark=mri] coordinates {(952536, 3906904)};
          \addplot[smooth,gray!30,mark=text,text mark=mya] coordinates {(424065, 3680839)};
          \addplot[smooth,gray!30,mark=text,text mark=nld] coordinates {(777896, 3788517)};
          \addplot[smooth,lightgray,mark=text,text mark=nor] coordinates {(698944, 3227602)};
          \addplot[smooth,lightgray,mark=text,text mark=nor] coordinates {(703960, 3254839)};
          \addplot[smooth,gray!30,mark=text,text mark=plt] coordinates {(660713, 3811113)};
          \addplot[smooth,gray!30,mark=text,text mark=poh] coordinates {(978619, 4957090)};
          \addplot[smooth,lime,mark=text,text mark=por] coordinates {(702841, 3359456)};
          \addplot[smooth,lime,mark=text,text mark=por] coordinates {(715175, 3362080)};
          \addplot[smooth,lime,mark=text,text mark=por] coordinates {(694258, 3447363)};
          \addplot[smooth,gray!30,mark=text,text mark=qub] coordinates {(505687, 3937018)};
          \addplot[smooth,gray!30,mark=text,text mark=quh] coordinates {(524175, 4033651)};
          \addplot[smooth,gray!30,mark=text,text mark=quy] coordinates {(496839, 4137823)};
          \addplot[smooth,gray!30,mark=text,text mark=quz] coordinates {(481488, 3827058)};
          \addplot[smooth,gray!30,mark=text,text mark=ron] coordinates {(712957, 3468627)};
          \addplot[smooth,gray!30,mark=text,text mark=rus] coordinates {(584472, 2947813)};
          \addplot[smooth,gray!30,mark=text,text mark=som] coordinates {(685419, 3969567)};
          \addplot[smooth,gray!30,mark=text,text mark=tbz] coordinates {(660053, 3346812)};
          \addplot[smooth,gray!30,mark=text,text mark=tcw] coordinates {(609364, 3931760)};
          \addplot[smooth,gray!30,mark=text,text mark=tgl] coordinates {(801180, 4234368)};
          \addplot[smooth,gray!30,mark=text,text mark=tlh] coordinates {(756654, 3721903)};
          \addplot[smooth,gray!30,mark=text,text mark=tpi] coordinates {(1117575, 5079660)};
          \addplot[smooth,gray!30,mark=text,text mark=tpm] coordinates {(847952, 3482679)};
          \addplot[smooth,darkgray,mark=text,text mark=ukr] coordinates {(623019, 3022281)};
          \addplot[smooth,darkgray,mark=text,text mark=ukr] coordinates {(622653, 3022462)};
          \addplot[smooth,cyan,mark=text,text mark=vie] coordinates {(766881, 3405675)};
          \addplot[smooth,cyan,mark=text,text mark=vie] coordinates {(815496, 3392739)};
          \addplot[smooth,cyan,mark=text,text mark=vie] coordinates {(797813, 3322379)};
          \addplot[smooth,gray!30,mark=text,text mark=wal] coordinates {(612448, 3810849)};
          \addplot[smooth,gray!30,mark=text,text mark=wbm] coordinates {(967466, 4365805)};
          \addplot[smooth,gray!30,mark=text,text mark=xho] coordinates {(486375, 3324166)};
          \addplot[smooth,gray!30,mark=text,text mark=zom] coordinates {(739611, 3452945)};
      \end{axis}
    \end{tikzpicture}
  \end{adjustbox}
  \caption{Tokens and characters (as reported by \texttt{wc \mbox{-w}}/\texttt{-m}) of the 106 Bibles. Equal languages share a color, all others are shown in faint gray. Most Bibles have around 700k tokens and 3.6M characters; outliers like Mandarin Chinese (cmn) are not surprising.}
  \label{fig:bible-tokens-chars}
\end{figure}

The optimal solution that we find contains 25996 verses for 106 Bibles in 62 languages,\footnote{afr, aln, arb, arz, ayr, bba, ben, bqc, bul, cac, cak, ceb, ces, cmn, cnh, cym, dan, deu, ell, eng, epo, fin, fra, guj, gur, hat, hrv, hun, ind, ita, kek, kjb, lat, lit, mah, mam, mri, mya, nld, nor, plt, poh, por, qub, quh, quy, quz, ron, rus, som, tbz, tcw, tgl, tlh, tpi, tpm, ukr, vie, wal, wbm, xho, zom} spanning 13 language families.%
\footnote{22 Indo-European, 6 Niger-Congo, 6 Mayan, 6 Austronesian, 4 Sino-Tibetan, 4 Quechuan, 4 Afro-Asiatic, 2 Uralic, 2 Creoles, 2 Constructed languages, 2 Austro-Asiatic, 1 Totonacan, 1 Aymaran; we are reporting the first category on Ethnologue \citep{paul2009ethnologue} for all languages, manually fixing tlh $\mapsto$ Constructed language.}
The sizes of the selected Bible subsets are visualized for each Bible in \cref{fig:bible-tokens-chars} and in relation to other datasets in \cref{tab:dataset-sizes}.

We split them into train/dev/test by dividing the data into blocks of 30 paragraphs and then taking 5 sentences for the development and test set each, leaving the remainder for the training set. This way we ensure uniform division over books of the Bible and sizes of $\nicefrac{2}{3}$, $\nicefrac{1}{6}$, and $\nicefrac{1}{6}$, respectively.

\begin{table}
  \begin{adjustbox}{width=\linewidth}
    \begin{tabular}{|l|rrr|}
      \hline
      English corpus        &                   lines &                    words &                     chars \\
      \hline
      WikiText-103          &                 1809468 &                101880752 &                 543005627 \\
      Wikipedia \footnotesize \raisebox{.1em}{(\raisebox{.2em}{$\underset{\in \text{\texttt{[a-z ]*}}}{\text{\texttt{text8}}}$})}
                            &                       1 &                 17005207 &                 100000000 \\
      Europarl              &                   78169 &                  6411731 &                  37388604 \\
      WikiText-2            &                   44836 &                  2507005 &                  13378183 \\
      PTB                   &                   49199 &                  1036580 &                   5951345 \\
      62/106-parallel Bible &                   25996 &               $\sim$700000 &               $\sim$3600000 \\
      \hline
    \end{tabular}
  \end{adjustbox}
  \caption{Sizes of various language modeling datasets, numbers estimated using \texttt{wc}.}
  \label{tab:dataset-sizes}
  \vspace{-\baselineskip}
\end{table}

\section{Detailed regression results}

\subsection{WALS}\label{app:detailed-results-wals}

We report the mean and sample standard deviation of language difficulties for languages that lie in the corresponding categories in \cref{tab:wals-correlations}:

\begin{table}[h!]
  \begin{adjustbox}{width=\linewidth}
    \begin{tabular}{l|ll}
      \toprule
      \textbf{26A} {\footnotesize (Inflectional Morphology)} & BPE                                            & chars                                          \\\midrule
      1 Little affixation (5)                                &           -0.0263 {\footnotesize ($\pm$ .034)} & \phantom{-}0.0131 {\footnotesize ($\pm$ .033)} \\
      2 Strongly suffixing (22)                              & \phantom{-}0.0037 {\footnotesize ($\pm$ .049)} &           -0.0145 {\footnotesize ($\pm$ .049)} \\
      3 Weakly suffixing (2)                                 & \phantom{-}0.0657 {\footnotesize ($\pm$ .007)} &           -0.0317 {\footnotesize ($\pm$ .074)} \\
      6 Strong prefixing (1)                                 & \phantom{-}0.1292                              &           -0.0057                              \\\toprule
      \textbf{81A} {\footnotesize (Order of S, O and V)}     & BPE                                            & chars                                          \\\midrule
      1 SOV (7)                                              & \phantom{-}0.0125 {\footnotesize ($\pm$ .106)} & \phantom{-}0.0029 {\footnotesize ($\pm$ .099)} \\
      2 SVO (18)                                             & \phantom{-}0.0139 {\footnotesize ($\pm$ .058)} &           -0.0252 {\footnotesize ($\pm$ .053)} \\
      3 VSO (5)                                              &           -0.0241 {\footnotesize ($\pm$ .041)} &           -0.0129 {\footnotesize ($\pm$ .089)} \\
      4 VOS (2)                                              & \phantom{-}0.0233 {\footnotesize ($\pm$ .026)} & \phantom{-}0.0353 {\footnotesize ($\pm$ .078)} \\
      7 No dominant order (4)                                & \phantom{-}0.0252 {\footnotesize ($\pm$ .059)} & \phantom{-}0.0206 {\footnotesize ($\pm$ .029)} \\
    \end{tabular}
  \end{adjustbox}
  \caption{Average difficulty for languages with certain WALS features (with number of languages).}
  \label{tab:wals-correlations}
\end{table}

\subsection{Raw character sequence length}\label{app:detailed-results-charseq}

We report correlation measures and significance values when regressing on raw character sequence length in \cref{tab:charseq-correlations}:

\begin{table}[h!]
  \begin{adjustbox}{width=\linewidth}
    \begin{tabular}{r|l|ll|ll}
                                &           & \multicolumn{2}{|c|}{BPE} & \multicolumn{2}{|c}{char} \\
                        dataset & statistic & $\rho$ & p     & $\rho$ & p       \\\toprule
      \multirow{2}{*}{Europarl} & Pearson   & .509   & .0185 & \textbf{.621} & \textbf{.00264}  \\
                                & Spearman  & .423   & .0558 & \textbf{.560} & \textbf{.00832}  \\\midrule
      \multirow{2}{*}{Bibles}   & Pearson   & .015   & .917  & \textbf{.527} & \textbf{.000013} \\
                                & Spearman  & .014   & .915  & \textbf{.434} & \textbf{.000481} \\
    \end{tabular}
  \end{adjustbox}
  \caption{Correlations and significances when regressing on raw character sequence length.  Significant correlations are boldfaced.}
  \label{tab:charseq-correlations}
\end{table}

\subsection{Raw word inventory}\label{app:detailed-results-wordinventory}

We report correlation measures and significance values when regressing on the size of the raw word inventory in \cref{tab:wordinventory-correlations}:

\begin{table}[h!]
  \begin{adjustbox}{width=\linewidth}
    \begin{tabular}{r|l|ll|ll}
                              &           & \multicolumn{2}{|c|}{BPE} & \multicolumn{2}{|c}{char} \\
                      dataset & statistic & $\rho$ & p     & $\rho$ & p    \\\hline
    \multirow{2}{*}{Europarl} & Pearson   & .040   & .862  & .107   & .643 \\
                              & Spearman  & .005   & .982  & .008   & .973 \\\hline
    \multirow{2}{*}{Bibles}   & Pearson   & \textbf{.742} & \textbf{8e-12} & .034  & .792 \\
                              & Spearman  & \textbf{.751} & \textbf{3e-12} & -.025 & .851 \\
    \end{tabular}
  \end{adjustbox}
  \caption{Correlations and significances when regressing on the size of the raw word inventory.}
  \label{tab:wordinventory-correlations}
\end{table}

\end{document}